# Autonomous Advanced Aerial Mobility – An End-to-end Autonomy Framework for UAVs and Beyond

**Sakshi Mishra[1], Senior Member, IEEE, and Praveen Palanisamy[1], Senior Member, IEEE**

[1]Autonomous Systems Group, Project AirSim, Microsoft Business AI + Research, Redmond, WA 98052

Corresponding author: Sakshi Mishra (e-mail: sakshi.m@outlook.com)

**ABSTRACT** Developing aerial robots that can both safely navigate and execute assigned mission without any human intervention – i.e., fully autonomous aerial mobility of passengers and goods – is the larger vision that guides the research, design, and development efforts in the aerial autonomy space. However, it is highly challenging to concurrently operationalize all types of aerial vehicles that are operating fully autonomously sharing the airspace. Full autonomy of the aerial transportation sector includes several aspects, such as design of the technology that powers the vehicles, operations of multi-agent fleets, and process of certification that meets stringent safety requirements of aviation sector. Thereby, Autonomous Advanced Aerial Mobility is still a vague term and its consequences for researchers and professionals are ambiguous. To address this gap, we present a comprehensive perspective on the emerging field of autonomous advanced aerial mobility, which involves the use of unmanned aerial vehicles (UAVs) and electric vertical takeoff and landing (eVTOL) aircraft for various applications, such as urban air mobility, package delivery, and surveillance. The article proposes a scalable and extensible autonomy framework consisting of four main blocks: sensing, perception, planning, and controls. Furthermore, the article discusses the challenges and opportunities in multi-agent fleet operations and management, as well as the testing, validation, and certification aspects of autonomous aerial systems. Finally, the article explores the potential of monolithic models for aerial autonomy and analyzes their advantages and limitations. The perspective aims to provide a holistic picture of the autonomous advanced aerial mobility field and its future directions.

## I. INTRODUCTION

Today, we are witnessing a paradigm shift in the automotive industry, with the entire century-old industry shifting toward electrification and increasing levels of connectivity and autonomy. The four main drivers of this disruption are: (A) Electrification - need for sustainability and reduced carbon footprint of transportation sector (B) Autonomy - need for safety, affordability, scale (C) On-demand/personalization – need for personalized, affordable, on-demand, fast transport for people & goods. (D) Connectivity – rise of 5G[1]. This revolution has affected several aspects, including the business models of industry players, from century-old Original Equipment Manufacturers (OEMs) to insurance agencies, including automotive material suppliers (e.g., batteries), tier I, II, III suppliers, chipmakers, etc. to rethink their approach in the era of electrification, connectivity, and AI. In the aerial domain, the electrification revolution is just getting started [1] with a handful of companies building eVTOLs[2] [2]; and starting to build and deploy towards making air taxis a reality [3], among other applications, to create new industry segments in the form of Advanced Air Mobility (AAM), Urban Air Mobility (UAM), package delivery, inspection, etc. eHang, recently, has become the world's first eVTOL maker to be awarded the aircraft type certificate for its autonomous vehicle, the EH216-S [4]. In the United States alone, the AAM market is estimated to reach US$115 billion annually by 2035 [5].

### A. BACKGROUND

A myriad of such applications within the aviation industry, from small manned to unmanned aerial vehicles in both military and civilian sectors [6], are advancing using both automation and autonomy[3]. These technologies hold the promise of making aircraft easier to fly and improving safety in the traditional aviation sector. On the other hand, their promise to the new market segments is to relieve the traditional one-to-one piloting of aerial vehicles, the financial cost of which is a prohibitive factor for scaling, in favor of the potential of fleet-level management.

[1] 5G is a key infrastructural element enabling the Unmanned Aircraft Systems (UAS) Traffic Management (UTM) segment.

[2] An eVTOL is an aircraft capable of taking off, hovering, and landing vertically thanks to an electric propulsion system.

Additionally, with the introduction of the new aviation segments such as UAM, AAM, etc. the fundamental nature of aerial mobility in urban settings is poised to become a more personalized experience for consumers – leading to high volume operations within shared airspace [7]. To accommodate high volume of mobility under UAM segment, high density flight operations with geographically constrained and densely populated areas will need to be conducted in the airspace that's currently unavailable to commercial jet aircraft systems. With such developments, autonomous operations become all the more significant in realizing robust operational control over all aspects of aerial mobility including route planning, fleet management, spacing, battery-charging/energy-management optimization, and more. Therefore, increased autonomy is a fundamental facilitator for the viability and commercial scale growth of the newer market segments including UAM, delivery, and inspection applications. More specifically, AI is a key enabler in several components spread across all these segments with varying levels of autonomous operations.

### B. RELEVANCE AND LITERATURE REVIEW

It is imperative that operations, *at scale*, in the shared airspace– for old and new market segments alike – require autonomy. Today, most of the autonomous system solutions are based on AI (deep learning and deep reinforcement learning to be specific). For the autonomous aerial mobility domain, majority of the sub-tasks for enabling autonomous operations (perception [8], scene-understanding [9], localization [10], mapping [11], planning, and control [12]) can be accomplished more effectively using AI-based methods when compared to traditional computer vision, robotics, and controls methods.

No doubt that software technologies (simulation, machine learning, and other AI techniques) are going to play a pivotal role in bringing the vision of autonomous aerial mobility to fruition. In the most abstract form, the needed software technologies and tools can be classified into three major branches: i) Simulation; ii) Data; and iii) Autonomy. 'Simulation' pertains to high-fidelity simulation for aerial vehicles and environments where they may fly. 'Data' encompasses synthetic data generation, processing, and curation capabilities along with the ability to bring sensed/measured data from the real world to simulation world (we term this 'Dataverse' in this article). 'Autonomy' refers to AI-enabled software that supports a wide variety of applications in perception, scene-understanding, planning, and eventually controls for aerial vehicles. Suites of such software programs, operating in tandem, are what's needed to enable autonomous aerial mobility applications *at-scale* ranging from UAM to various other UAVs.

As can be seen from the intersectionality Venn diagram in Fig 1, this problem is multidisciplinary in nature and requires components from different disciplines to be put together to build the technology which can safely enable autonomous advanced aerial mobility. Moreover, these components must be compatible with each other and in the end synergize with each other to meet the stringent safety standards of aviation field.

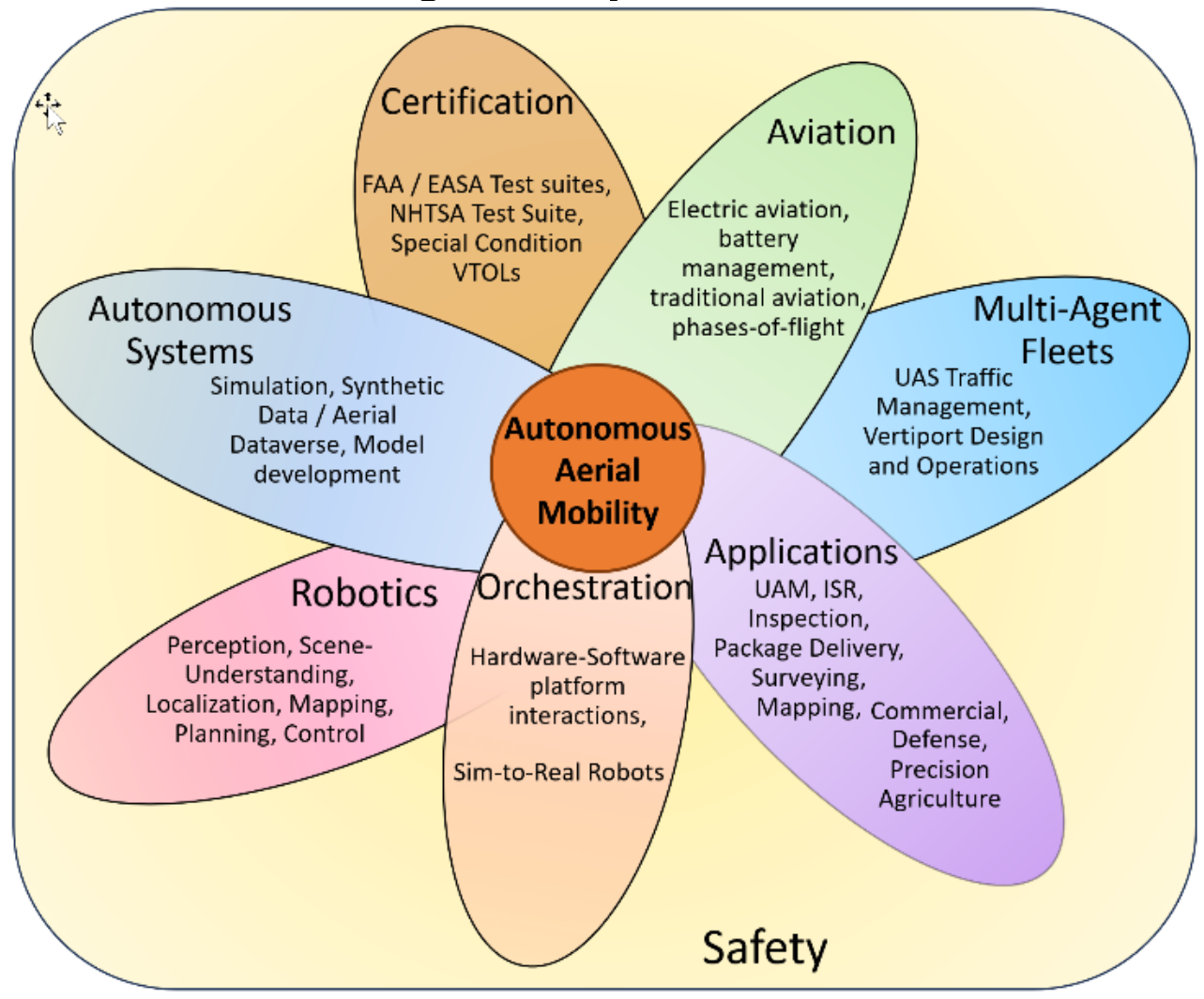

**FIGURE 1 Advanced Autonomous Aerial Mobility – Presented in a holistic and interdisciplinary details in this perspective.**

Additionally, developing and deploying autonomous AAM systems poses significant technical and operational challenges, such as ensuring safety, robustness, scalability, and efficiency in complex and dynamic environments [13]. Despite the recent advances in artificial intelligence (AI), robotics, and simulation technologies, the literature on UAVs doesn't comprehensively present a holistic picture of how to design autonomy blocks and technology stack for aerial mobility that's flexible and adaptable to different form-factor and applications, with multi-fleet operations considerations, and furthermore, considering the certification aspects [14] [15] [16] [17] [18] [19].

### C. CONTRIBUTIONS OF THIS WORK

In this work, we address the above-mentioned gaps in the literature. This is a *Perspective article* that presents a holistic qualitative approach to Autonomous Advanced Aerial Mobility. More specifically, the contributions of this work are as follows:

- We put-forth our perspective on how the *autonomous aerial mobility technology* field will evolve and converge into a modular framework that gives aerial robots the ability to 'see', 'understand', 'decide', and 'move' using artificial intelligence based algorithms and methods.

[3] The difference between autonomy and automation is discussed in Section II.A.

- To put our proposed perspective in a balanced light, we also discuss the notion of 'monolithic deep learning for autonomous aerial vehicles' – which refer to the paradigm that one single end-to-end model should be trained to execute all four major tasks as opposed to the modular framework that we propose in this perspective.
- We propose a scalable and extensible Autonomy Block stack for Advanced Aerial Mobility with detailed explanation of the functionalities and technical underpinnings of various blocks.
- For each topic discussed in this perspective, we canvass various key research works that have been published on the sub-topics and sub-technologies that constituents the proposed framework.
- We bring together multi-faceted aspects of Autonomous Advanced Aerial Mobility field to present a holistic picture of the field including:
  - Role of simulation, synthetic data, and AI in enabling autonomy
  - Applications across domains, sectors, and scenarios
  - Multi-agent fleet operations and progress made by agencies such as Federal Aviation Administration (FAA) and EASA on orchestrating the operations of large-scale deployments of UAV fleets.
  - Testing, validation, and certification of these new AI-based technology from regulatory bodies.

It is important to note that this work is not a survey of the research works that have been published in the UAV field. Instead, this article presents a technical "perspective" on how the autonomous advanced aerial mobility field will evolve keeping autonomy technology and AI at center. Various sub-fields under this umbrella topic are explored in this work for bolstering the presented perspective with conceptual underpinnings of the interrelated topics. Moreover, monolithic approach is also discussed as a contrasting school-of-thought to put forth our modular framework in a balanced light.

### D. ARTICLE ORGANIZATION

Section 1 introduces the background, motivation, and contributions of this work on autonomous aerial mobility. Section 2 contextualizes the fundamentals of autonomy, automation, and AI for aerial vehicles and compares them with ground vehicles. Section 3 discusses the roles of simulation and synthetic data in advancing autonomous systems and presents the Aerial DataVerse toolchain. Section 4 proposes the Autonomy Blocks framework for advanced aerial mobility and explains its components and functionalities. Section 5 surveys various applications of autonomous aerial mobility across domains, sectors, and scenarios. Section 6 introduces multi-agent fleet operations and management and reviews the current progress and challenges in UTM. Section 7 focuses on benchmarking and validation aspects of AI models for autonomy blocks and discusses the safety and certification issues. Section 8 explores the potential of foundation models for aerial autonomy and analyzes their advantages and limitations. Section 9 concludes the article and provides the outlook on the Autonomous Aerial Mobility field.

## II. CONTEXTUALIZATION OF FUNDAMENTALS

### A. AUTOMATION VS AUTONOMY

The rapidly advancing aviation industry is proposing and developing intelligent systems and solutions for different phases of flights, different types of vehicles, various operating envelops, and different sizes of sub-systems within the vehicles. Consequently, the terms automation and autonomy are sometimes used interchangeably. Though somewhat related, these two terms signify two very different notions of human-intervention-independent task-execution by machines. The key differentiation between the two needs to be established before we go on to describe the proposed Autonomy Blocks framework for aerial mobility.

To date, there are no standard definitions differentiating automation and autonomy. However, the industry is converging to the following – Automation is a process performed without human assistance which typically runs within a well-defined set of parameters. Automated systems (or sub-systems) are very restricted in what tasks they can perform, are designated to accomplish a specific set of largely deterministic steps to achieve a limited set of pre-defined outcomes [20]. On the other hand, autonomy implies satisfactory performance under significant uncertainties in the environment and the ability to compensate for system failures (i.e., built-in software redundancies) without external intervention. Autonomous systems learn and adapt to dynamic environments.

An example of automation in the aviation domain is autopilot technology. These systems are programmatically designed to keep the aerial vehicle (most often commercial jets) level and headed in the right direction; however, departure from the usual operational envelop or circumstances requires a human pilot to take over and human supervision is needed all the while. On the other hand, a fully autonomous aerial vehicle (e.g., delivery drone) operates without human assistance in a dynamically changing environment. In this scenario, the starting point and the end point are fixed, the route has been pre-planned, but the on-board vehicle intelligence needs to dynamically respond to wind-gusts forcing it to deviate from its pre-planned waypoints, and also determine the right landing spot by finding an obstacle-free zone on the driveway of the delivery location. In this case, the autonomously operating vehicle is making its own judgements and acting under uncertainty. To summarize, autonomous systems are akin to a living/constantly evolving (artificial) intelligence that can progressively take over the complete higher-level task (e.g. flying an eVTOL) by continuously learning from its environment. Automation, on the other hand, can only perform preprogrammed tasks that are repetitive in nature with little to no understanding of the changing environment around it.

Another way to understand autonomous systems is - a constantly evolving and dynamic process that is capable of perceiving the on-going changes in the real-world and responding appropriately in order to meet the objective of the process (in our case – navigating an aerial vehicle safely to its destination to complete the mission). There are four major components of this process – 1) Sense; 2) Perceive; 3) Plan (or decide); 4) Actuate. This view of autonomous system, with its components, is depicted in Fig 2. Advanced Aerial Mobility, as an autonomous system, is discussed in detail in Section 4.

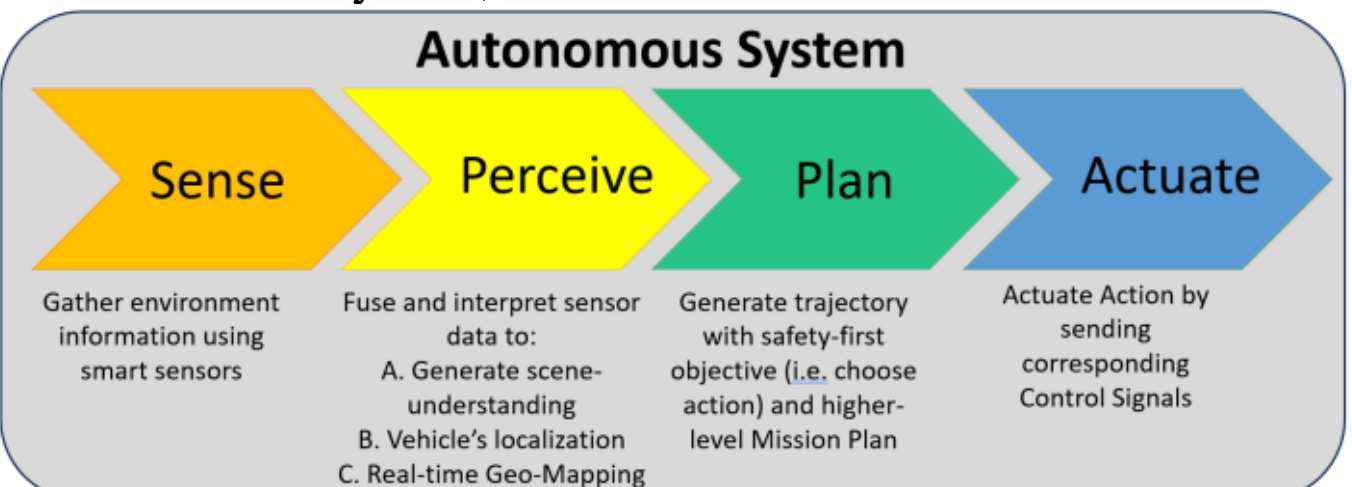

**FIGURE 2 Four Major Building Blocks of Autonomous Systems**

#### 1) *Rule-based vs Learning-based Approach*

Autonomous systems have learning-based methods at their core. Artificial neural networks (ANN) are universal function approximators; that is, it is possible to represent complex nonlinear behavior in a high-dimensional space using ANNs. A deep neural network is an ANN with multiple hidden layers and nodes cascaded between input and output layers. Deep neural networks are sophisticated neural networks that have been successfully applied to analyze data in many disciplines in the past several years such as computer vision, image recognition, automatic speech recognition, bioinformatics, finance, and natural language processing [21]. In general, traditional machine learning algorithms such as decision trees, Naïve Bayes classifiers, K Nearest Neighbors etc. are particularly task specific. However, deep learning networks are capable of learning intricate structures in large datasets, allowing them to generalize better to address all the scenarios – however non-linearly related – that are included in the training dataset. Additionally, deep learning algorithms typically do not require the type of extensive feature engineering that is required of other traditional machine learning methods[4].

Another class of learning-based approaches that have the potential to enable full autonomy is Deep Reinforcement Learning (DRL). DRL is a branch of machine learning that combines deep neural networks with reinforcement learning, a technique that learns from trial and error by interacting with an environment. DRL agents can learn complex and optimal policies for sequential decision-making problems, such as controlling an aerial vehicle, without requiring explicit supervision or prior knowledge. DRL has been successfully applied to various domains, such as robotics, games, and self-driving cars [22] [23] [24].

Rule-based approaches, on the other hand, do not have any generalization capabilities. Furthermore, closed form analytical equations based models do not account for the changes in the environment and rigidly follow the constructs with which they are written by humans. There is no "continuous" improvement process that allows rule-based approaches to evolve to cater to higher and higher machine intelligence needs. For applications where the number of scenarios and variations are vast – it is extremely challenging, inefficient, and practically close to impossible to create intelligent systems using rule-based programs that can sense, perceive, understand, and act in real-time for dynamically changing environments.

### B. *AUTONOMY FOR 2D vs 3D TRANSPORTATION*

Given the advances that have been made in the driverless ground transportation industry, the comparison of the technological parallels between the two mediums of mobility is often drawn. We summarize the main differences between the two (2D and 3D transportation) below:

- The degrees of freedom in case of aerial transportation (x, y, z axes and roll, pitch, yaw) is twice as many as ground transportation (x,y axes, and orientation) – so for fully autonomous navigation, this is a relatively harder problem to solve when 6 variables are to be determined by the algorithms that are solving multi-objective optimization problem, with even more stringent safety criterion, where the objectives and constraints include:
    - minimize fuel/battery consumption to maximize flight-time/distance covered,
    - minimize distance by picking shortest possible trajectory (including minimizing altitude changes),
    - detect and navigate around the obstacles that could show up from various directions such as top, bottom, forward, side-ways (i.e. it has more *collision-potential*),
    - keep the vehicle stabilized in the air while navigating around dynamic obstacles, wind gusts, etc.
    - maximize safety (i.e. avoid operational failure at all costs)
- Autonomous aerial vehicle operation is essentially a complex controls problem where the onboard intelligence must sense, perceive, process, and give control commands to the rotors all the while maintaining the base level stability of vehicle in the air to counter the thrust, drag, list, and weight forces (form-factors with hovering capabilities have slightly different sets of constraints to operate than the form-factors without hovering functionality). The control commands to the rotors need to be managed within a *matter of milliseconds to seconds* while counteracting these forces, localizing, mapping, path-planned, detecting and avoiding obstacles, etc. Ground transportation, on the other hand, has a smaller number of sheer physical forces to counter as they are stuck to the ground throughout the process. This makes 3D mobility a harder problem to tackle.

[4] As the autonomous driving industry is striving to get to Level III automation certified for deployment, there are concerns in the AI-community about Deep Learning's ability to deliver Level V (fully autonomous) vehicles. It is asserted that deep learning can only interpolate. Deep neural networks extract patterns from data, but they don't develop causal models of their environment. This requires the training dataset to cover all "edge-cases" or different nuances of the problem that the deep learning model will encounters upon deployment.

- The onboard sensing and compute hardware of aerial vehicles is tightly intertwined with the size, weight, and power (SWaP) constraints of the aerial vehicle [25]. The type of sensors (also referred to as optical payload) and the size of compute that can be installed for an application-specific UAV is dependent on various factors such as SWaP limitations, development and unit cost reduction targets (together termed as SWaP-c constraints). Therefore, bringing higher and higher levels of intelligence in the aerial autonomy domain — when compared to the driverless ground vehicles domain — is a harder problem to solve.
- Safety considerations: the aviation sector is known for its one of the most rigorous safety standards across different industries, which again makes it a more involved problem to solve compared to the ground transportation sector.
- The UAM segment has vehicles with different form factors (from small inspection drones to air-taxi carrying passengers) that vary in their operating speed ranges and maneuvering capabilities. When these vehicles share the same airspace, creating autonomous navigation capabilities with fast detect-and-avoid capabilities is a necessity. To this end, the research and regulatory communities are actively working on designing airspace for dense UAM future [26]. In the case of driverless ground vehicles, the navigation route infrastructure and driving rules are pre-established and by design less prone to collisions than vast airspace for 3D transportation where lanes/speed-operating-zones/ are not [yet] established[5].
- The aerial domain has more environmental disturbance, more pronounced in the case of smaller form factor vehicles. For example, hail or a gust of air easily gets the drone destabilized or forcefully stray it from the planned trajectory.
- Driving only happens on roads and every inch of the road network has been high-definition (HD) mapped [27]. HD mapping every inch of the volumetric air space is an extremely challenging (i.e. compute intensive) problem. Due to the lack of HD-mapped space, more sophisticated methods and/or combination of algorithms are needed to solve localization, mapping, and path planning problems in the aerial domain.

There is one aspect however, that makes ground transportation's autonomous operations more challenging than aerial transportation is – Pedestrians. Pedestrians are not on the way for the majority of aerial vehicles (*exception* - delivery drones that get closer to the ground during a drop-off in front of public buildings/houses). Autonomous ground vehicles operating in urban environments must predict pedestrian behavior which is highly stochastic in nature. This increases the complexity of decision-making for autonomous ground transportation.

### *C. LEVELS OF AUTONOMY*

Advancements in the aviation industry – both the vehicle types/designs and the aviation ecosystem in which they operate – are shaped and guided by complex trades among multi-disciplinary technological innovations in engineering domains, economics, baseline technical requirements, and sustainability concerns, all built on a foundation of *safety*. The level of autonomy, therefore, is determined by a complex set of factors such as - how much is the human pilot (remote or on-board) involved? How much is controlled by AI-based intelligence? How much control does the traditional/rude-based automation retain? How complex is the environment? How can inter-agent interactions be modeled? At what level of trust? What's the level of redundancy needed before machine intelligence can take over a significant portion of decision-making? For the ground transportation sector, the Society of Automotive Engineers (SAE) has classified autonomous driving into six levels in the standard published in 2014 SAE J3016 [28]. Comprehensive and widely/universally accepted aerial autonomy standards are in making but it is highly likely that the progression of the autonomy level in the future aviation specific standards will be similar to autonomous driving.

Towards the second half of autonomy levels on the spectrum, human pilot's ability to override the system decision-making and action-sequence is a key transition phase. European Union Aviation Safety Agency (EASA)'s concept paper [29] highlights three distinctive phases of machine intelligence design where overriding capabilities of the human pilot are progressively phased out: i) Overseen and overridable[6]; ii) Overridable[7]; iii) Non-overridable[8]. The paper also classifies the five levels of autonomy as: i) Human Augmentation; ii) Human assistance; iii) Human-AI collaboration (overseen and overridable); iv) More autonomous AI (overridable); v) Fully Autonomous AI (non-overridable).

Human Machine Interaction (HMI) is the study of how humans and machines interact in complex systems, such as aviation. HMI involves the design, evaluation and implementation of interfaces that facilitate the communication, coordination and control between humans and machines. HMI is especially important in aviation, where pilots, air traffic controllers, maintenance crews and other stakeholders rely on various technologies to ensure safe and efficient flight

[5] NASA's ongoing effort in ATM: What is the Air Traffic Management eXploration? | NASA

[6] [29] - "capability for the human to closely monitor the functions allocated to the AI-based system (every decision-making and action implementation), with the ability to intervene in every decision-making and/or action implementation of the AI-based system."

[7] [29] - "capability for the human to supervise the operations of the AI-based system (some decision-making and some action implementation), with the ability to override the authority of the AI-based system (some decision-making and some action implementation) when it is necessary to ensure safety and security of the operations (e.g., upon alerting)."

[8] [29] - "human has no capability to override the AI-based system's operations"

operations. HMI in aviation aims to optimize the performance, workload, situation awareness and error management of human operators, while also enhancing the reliability, usability and adaptability of machines. HMI takes on added importance in the Autonomous Operations realm as the overall field evolves to integrate higher levels of autonomy in the aerial vehicles. Balancing the roles and responsibilities of humans and machines; ensuring the compatibility and interoperability of different systems; and addressing the ethical, legal and social implications of HMI are the challenges that need to be addressed as the technical advancements unfold on autonomy front.

### D. KEY DIFFERENCES BETWEEN AAM AND TRADITIONAL AVIATION

Advanced Aerial Mobility (AAM) is a term that describes a new era of air transportation that uses highly automated and electric aircraft, such as air taxis or eVTOL aircraft. AAM aims to provide safe, accessible, affordable and sustainable air travel for passengers and cargo in urban and rural settings. AAM aircraft can perform various missions, such as package delivery, emergency response, aerial observation and personal transportation. Advanced Air Mobility AAM differs from traditional aviation in several keyways:

- Mission Distances: AAM typically involves shorter mission distances compared to traditional aviation, which usually involves flights of greater distances.
- Aircraft Technology: AAM involves the use of new airborne technologies such as electric and hybrid aircraft for urban, suburban, and rural operations. Traditional aviation primarily uses fuel-powered aircraft.
- Operational Environment: AAM aims to transport people and goods to locations not traditionally served by current modes of air transportation, including both rural and more challenging and complex urban environments. Traditional aviation mainly operates between established airports.
- Navigation and Timing: AAM missions would likely rely on precise navigation and timing through three-dimensional corridors of uncontrolled airspace. Traditional aviation operates in controlled airspace with established air traffic control procedures.
- Aircraft Types: AAM includes small drones, electric aircraft, and automated air traffic management among other technologies. Traditional aviation primarily involves manned aircraft.
- Safety, Sustainability, Affordability, Accessibility: These are highlighted as key features of AAM missions.

In essence, AAM represents a transformative approach to air travel that leverages new technologies and operational concepts to expand the reach and efficiency of aviation.

### E. KEY TERMINOLOGIES IN AAM SYSTEM

In this section we discuss a few key terminologies in the aviation and AAM ecosystem that are relevant to the presented work on Autonomous Aerial Mobility.

#### 1) ConOps

In aviation, Concept of Operations (ConOps) is a document that describes a proposed system concept and how that concept would be operated in an intended environment. The user community develops ConOps to communicate the vision for the operational system to the acquisition and developer community. It is designed to give an overall picture of an operation and facilitate a common understanding of a future system to help develop operational and system-level requirements. One of the widely utilized examples of ConOps in aviation include NASA's Concept of Operations Annotated Outline [30]. ConOps for Unmanned Aircraft Systems (UAS) are in active development. Section VI.A discusses this in detail.

#### 2) Visual Flight Rules (VFR) and Instrument Flight Rules (IFR)

Visual Flight Rules (VFR) and Instrument Flight Rules (IFR) are two different sets of rules that apply when flying an aircraft. VFR refers to the rules and regulations of operating an aircraft in weather conditions that are good enough for the pilots to see the horizon and where the aircraft is going[1]. Pilots cannot fly using VFR if they are flying through clouds or within the defined clearances of them as they need to be able to see other aircraft[2]. The weather must be better than the VFR weather minima. Air Traffic Control (ATC) is not necessarily responsible for keeping planes that are flying VFR separated, though services such as flight following are available depending on the region.

On the other hand, IFR refers to the rules and regulations established by the FAA to govern flights under conditions in which flight, by outside visual reference, is not safe. This means that IFR only refers to flight done using aircraft instruments instead of depending solely on the visual of the pilot outside the aircraft. Instruments are used in low visibility scenarios such as bad weather or nighttime. IFR gives an aircraft the authority to operate under Instrument Meteorological Conditions (IMC) which means that an aircraft will be allowed to fly in any weather conditions less than the VMC (visual) but is still borderline safe.

Whether pilots fly VFR or IFR will depend partially on the weather conditions, the route of the flight, and other variables. All pilots flying in Class A airspace must have IFR and the pilots must be flying under IFR, regardless of the current weather conditions in the airspace.

#### 3) Flight Envelop and Operating Envelop

The flight envelope, also known as a performance envelope, refers to the design capabilities of an aircraft. It is typically expressed in terms of airspeed and load factor. The purpose of a flight envelope is to define the operational limits for an aerial platform with respect to maximum speed and load factor given a particular atmospheric density. It is determined during the design phase, where engineers calculate limits for maximum speed, altitude, load factor, and maneuverability.

In terms of safety, the flight envelope is crucial because it ensures that the aircraft operates within its designed structural capabilities. This minimizes the risk of over-controlling, losing control, overstressing, or damaging the aircraft. If an aircraft operates outside its flight envelope, it may suffer damage. From a certification perspective, airworthiness certification verifies that a specific air vehicle can be safely maintained and operated within its described flight envelope. It shows that the air vehicle can safely attain, sustain, and terminate a flight in accordance with approved usage limits (range, speed, weight, altitude, safety).

The operating envelope (also known as the operational flight envelope), on the other hand, refers to the area inside the boundaries that limit the normal flight operations of an aircraft. It is important to recognize the fundamental difference between the manufacturer's certified limits (flight envelope) and the airline's operating limits (operating envelope). The certified envelope provided in the aircraft flight manual (AFM) represents the approved safe limits for the airplane. However, it is not intended for use in actual load planning. While the flight envelope describes the maximum capabilities of an aircraft as determined by its design, the operating envelope describes the practical operational limits under which an aircraft is typically flown.

In the context of AAM, the flight envelope and the operating envelope take on added significance. AAM involves the use of new vehicle technologies that redefine the scale and types of operations possible in airspace systems. To facilitate the safe large-scale deployment and acceptance of these new technologies, public and private institutions must work together to understand and define the flight envelopes and operating envelopes for these vehicles, especially when they are equipped with autonomous navigation capabilities.

## III. ROLES OF SIMULATION AND SYNTHETIC DATA IN ADVANCING AUTONOMOUS SYSTEMS

Developing and testing autonomous systems in the real world can be challenging, time-consuming, and expensive. Simulation and synthetic data have emerged as essential tools in bringing autonomous systems to life by enabling efficient development, testing, and validation.

### *A. SIMULATION*

Simulation is a powerful tool for developing autonomous systems like unmanned vehicles. It allows researchers and engineers to test and evaluate different scenarios, algorithms, and designs virtually without risking the safety of the system or the environment in the real world. Simulation can also reduce the cost and time of development by enabling faster iterations and feedback loops. Furthermore, it also helps in validating the performance and robustness of the system under various conditions and uncertainties. By using simulation, autonomous systems can be improved, optimized, and safety-tested before deployment in the real world.

The proposed Autonomy Blocks framework leverages the AirSim simulator [31] to collect annotated training data on a large scale, encompassing millions of data points across a variety of environmental conditions and autonomy scenarios. The process of collecting the desired annotated training dataset is often sequential and follows the given steps:

- Digital 3D assets: Acquire or create them. In case of aerial vehicle simulation, this includes simulating physics, vehicle dynamics, controllers, and batteries.
- Scene Generation: Generate desired scene with 3D asset placement. Prerequisite here is access to or creation of base environments. For example, indoor – warehouse, outdoors – airports /fields.
- Procedural Scenario Generation: Design and implement various scenarios that simulate different real-world situations for training purposes. This includes inducing faults, adverse conditions, terrain, corner scenarios, and perturbations. Additionally, simulate weather conditions across flight envelope to account for environmental factors.
- Batch Generation of Annotated Synthetic data: By instantiating one or multiple simulation instances, generate and annotate large batches of synthetic data by incorporating the 3D assets and procedural scenarios, ensuring a diverse, pragmatic, and comprehensive training dataset.

### *B. AERIAL DATAVERSE: MULTI-MODAL, QUERYABLE SYNTHETIC TRAINING DATA*

In this section we trace fuel that powers the proposed Autonomy Blocks framework. The saying goes – "data is the new oil", authors' amendment to this proverb– "raw oil can't be pumped into engine, it needs economical extraction and effective recovery to become useful". Simulation or digital twin of the airspace, given its current state of development, captures a subset of the complexity of the real world. Moreover, a lot of manual engineering effort is required to create each new dataset, as described in the previous subsection. More importantly perhaps, it is for the systems designers/architect to carefully engineer the virtual/training scenarios that also encompass the edge-cases that keep the advanced autonomous solutions from getting certified on safety grounds. The sheer data-engineering resources needed to create such close-to-real-life-edge-cases scenario is often a prohibiting factor from a Machine Learning Operations (MLOps) point of view. Therefore, to facilitate the training of various AI models that form the backbone of autonomy blocks with proper datasets, a bridge is needed between the simulation and autonomy worlds.

To solve this *extraction* and *recovery* problem in the context of aerial mobility, we propose Aerial DataVerse – the fundamental building block of the presented overall autonomy framework. It is comprised of a toolchain for query-able data generation, curation, and data augmentation using generative models and AI technologies (NeRF [32], GANs [33], etc.). The aerial DataVerse is designed to collect billions of *eventful* samples to create high-entropy surprise datasets and simulation environments using AI actors. In the case of aviation domain, the cost of collecting eventful (entropy-rich) data in the real-world is exorbitant. In the aerial simulation platform, collecting high-entropy eventful samples is challenging too. Defining or scripting rare-events, unusual trajectories, and atypical human-machine collaboration experiences is (yet) not feasible without considerable amount of manual engineering effort using existing simulation tools and platforms. The proposed aerial DataVerse can help enable the following functionalities for creating datasets that closely emulate real-world conditions and do not need repetitive effort-intensive manual simulation engineering:

1) Query-able data generation

With the goal of creating datasets that can capture edge-cases that are anticipated by human intelligence before the autonomous system is put into production, query-able data generation capability is added to the proposed DataVerse. The module is designed to hot infer curated datasets based on the query entered (behind the scenes the module takes care of making necessary changes in all the config files for the simulation platform and data-collection pipeline as well as activating the AI actors).
Sample query: "Multiple-operator aerial traffic scenarios during thunderstorms"

- **Multiple operator:** More than one UAM operators involved ➔ Multiple aircraft models and types involved. Single-passenger to multi-passenger aircraft with different airframes (VTOLs/multicopters/planes). AI actors in this DataVerse to fly the different air taxis with different policies to simulate realistic conflicting traffic.
- **Aerial traffic scenarios during thunderstorms:** Conflicting trajectories due to re-routes based on the inclement weather (thunderstorm) advisory. Domain randomizations to account for and replicate thunderstorm scenarios, wind dynamics, sensor aberrations etc.

2) Domain Randomization

The "domain" here pertains collectively to the {environment, scenario, sensor parameters} and "randomizing" it would yield a system capable of generating datasets that are rich in domain knowledge spanning the spectrum of possible domains including the real (physical) domain.

3) Data Augmentation

Data augmentation includes post-processing of datasets to add variations and diversity to increase the information content in the datasets and sim sets thereby increasing the resiliency and robustness of the system utilizing the data. The applicable augmentations include crops, perspective transforms (tilt, skew, warp left/right/forward/backward), size, rotation, shears, random masks/erasures etc.

In addition to simulation generated data augmentation, the real-world data gathered by the sensing infrastructure on the deployed drone fleet will bring enormous value for continually improving the information entropy in the datasets and sim sets. As the adoption of UAVs grows, with (Azure-IoT powered) runtime, refinement telemetry, a set of "feedback signals" from the (real-world) deployments could be leveraged to help refine, improve, and finer-tune the pre-trained models embedded in the proposed Autonomy Blocks framework.

## IV. AUTONOMY BLOCKS – THE TECHNOLOGY THAT INFUSES ON-BOARD INTELLIGENCE

Developing aerial robots that can both safely *navigate* and execute assigned *mission* (task) without any human intervention – i.e., fully autonomous aerial mobility – is the larger vision that guides the research, design, and development efforts in the *aerial autonomy* space.

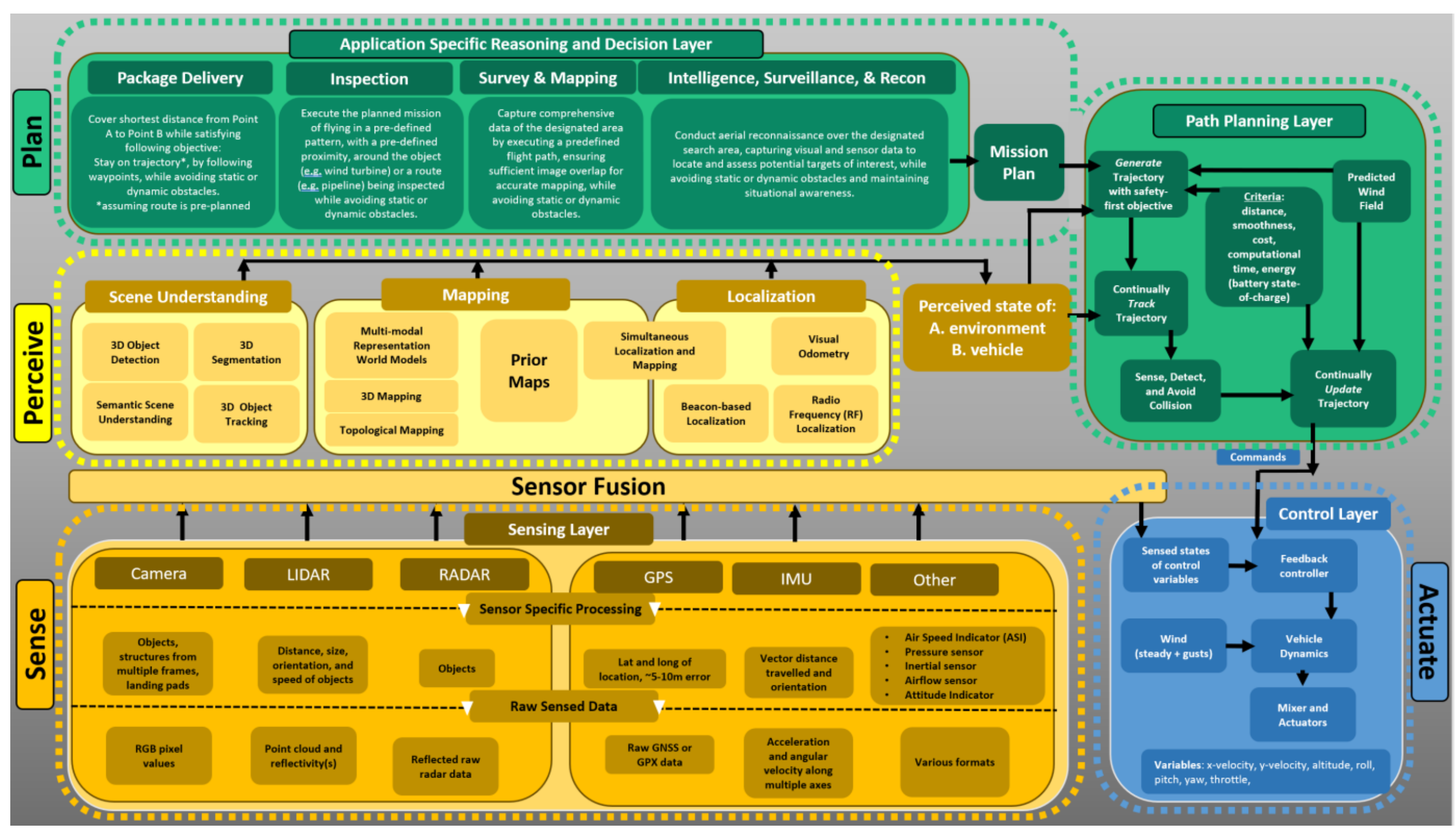

FIGURE 3 Autonomy Blocks Framework for Aerial Mobility

For some applications specific scenarios such as inspection and 3D-mapping of physical assets, autonomously navigating small scale UAVs have been tested, certified, and rolled out as commercial products in recent years, Skydio [34] being one of the prime examples. As the technology matures further, developing autonomy in all shapes, sizes, form-factors, and types of UAVs in a holistic way such that it supports their intertwined operations in shared airspace is a significant challenge that industry, researchers, regulatory bodies, and governments are grappling with.

A broad definition of Autonomous Systems with its four main building functional blocks was laid out in Section II.A. In this section, we present a detailed architecture of Autonomy Blocks designed for AAM applications. Additionally, we map various autonomy blocks with various phases of flight to further contextualize the modular design of the proposed autonomy blocks framework.

### A. AUTONOMOUS OPERATIONS – A CONGLOMERATION OF ROBOTICS PROBLEMS

Autonomous Operations in Advanced Aerial Mobility represents a multifaceted challenge that encompasses a conglomeration of robotics problems. To enable unmanned aerial vehicles (UAVs) or urban air mobility (UAM) systems to navigate and operate autonomously in complex environments, a comprehensive solution is required. This includes addressing issues related to sensing, such as developing robust and high-precision perception systems to detect and understand the surrounding environment. Additionally, conducting Localization becomes crucial for accurate positioning, often requiring advanced techniques like simultaneous localization and mapping (SLAM) to create real-time maps. Moreover, control algorithms must be finely tuned to ensure safe and precise maneuvering of these aerial vehicles amidst dynamic and unpredictable conditions. These elements, among others, must seamlessly integrate to achieve the overarching goal of enabling safe and efficient autonomous operations in the realm of advanced aerial mobility.

As can be inferred from the above discussion, autonomous AAM certainly requires more than just vehicle dynamics. Other elements such as intricate environmental conditions (weather, time-of-day, wind speed and direction, flying altitude), various sensor configurations and layouts, and airspace rules and traffic must also be taken into account for building the software systems to facilitate safe and effective operations. Fig 3 shows the proposed Autonomy Blocks framework. The four major blocks (sensing, perception, planning, and actuation) are further broken-down into sub-modules depicting the underlying sensing mechanisms, data processing modules, data flow patterns, and robotics algorithms that accomplish various sub-task for enabling aerial vehicle's navigation in its specific environment.

#### 1) Sensing

Sensing refers to the process of collecting data or information from the environment using various sensors and sensory technologies. Sensing involves capturing raw data from the surrounding environment, such as visual imagery, distance measurements, speed, altitude, and other relevant information. In the context of aerial mobility, sensors are essential for understanding the aircraft's surroundings (i.e. environment) and its own state (vehicle's dynamics, configurations, etc.), helping it gather real-time data. These sensors can include cameras, LiDAR (Light Detection and Ranging), radar, GPS (Global Positioning System), IMUs (Inertial Measurement Units), and more. Main types of sensors employed in AAM applications are listed and described below:

- **RGB Cameras**: RGB imaging devices operate within the visible spectrum and find applications in tasks such as object recognition, obstacle avoidance, and navigation.
- **Stereo Cameras**: Utilizing multiple lenses, stereo cameras capture images from distinct perspectives, facilitating the computation of depth information. Their utility spans tasks such as obstacle avoidance, navigation, and mapping.
- **Stereo Omnidirectional Cameras**: These cameras possess omnidirectional capabilities while maintaining stereo vision, enhancing their utility in various applications.
- **Monocular Cameras**: Monocular cameras provide singular perspective imagery and are employed in scenarios where depth perception is not a primary requirement.
- **Monocular Omnidirectional Cameras**: Combining monocular vision with omnidirectional capabilities, these cameras offer unique advantages in specialized applications.
- **Infrared Cameras**: Infrared cameras are designed to capture images in the infrared spectrum, serving functions like heat signature detection, obstacle avoidance, and navigation.
- **Thermal Sensors**: Thermal cameras create images predicated on temperature differentials, and are instrumental in tasks such as heat signature detection, obstacle avoidance, and navigation.
- **GPS**: Global Positioning System (GPS) is a satellite-based navigation system that provides accurate positioning information over a wide area. While GPS is widely used for outdoor localization, it may have limitations in terms of accuracy and availability in dense urban environments or indoor settings.
- **IMU**: An Inertial Measurement Unit (IMU) is an electronic device that measures and reports a body's specific force, angular rate, and sometimes the orientation of the body, using a combination of accelerometers, gyroscopes, and sometimes magnetometers. IMU measures the acceleration, angular velocity, and sometimes the magnetic field of a host device in three-dimensional space. The raw data from these sensors can be processed to determine the orientation and movement of the device. However, an IMU alone cannot provide the absolute location of a device in terms of latitude and longitude. Instead, it provides relative position data from a known starting point. To obtain the absolute location of a device, an IMU is typically used in conjunction with other sensors such as GPS or other external references. The output format of an IMU can vary depending on the specific device and its intended use, but it typically includes data on acceleration, angular velocity, and orientation in three-dimensional space. This data can be used to track the movement and orientation of the device over time. IMUs tend to accumulate errors over time and require correction from other sensors.

Recent developments allow for the production of IMU-enabled GPS devices. An IMU allows a GPS receiver to work when GPS-signals are unavailable, such as in tunnels, inside buildings, or when electronic interference is present. IMUs contain sensors such as accelerometers, gyroscopes, and magnetometers. Each tool in an IMU is used to capture different data types:

- Accelerometer: measures velocity and acceleration
- Gyroscope: measures rotation and rotational rate
- Magnetometer: establishes cardinal direction (directional heading)

In summary, IMUs are sensing devices that incorporate at least two (and often three) types of sensors to measure a host device's location in three-dimensional space. They are a valuable supplement to GPS or other navigational technologies.

- **LiDAR or depth sensor**: LiDAR (Light Detection and Ranging) cameras use laser light to measure distances and create 3D maps of the environment. By creating a detailed map of the surroundings and comparing it to real-time measurements, a vehicle can estimate its position relative to the map. They are used for tasks such as obstacle avoidance, navigation, and mapping.
- **Radar**: Radar (Radio Detection and Ranging) is a technology that employs radio waves to detect objects, measure their distances, and track their movements within the surrounding environment. By emitting radio frequency signals and analyzing the reflected signals (echoes), radar systems can create a comprehensive understanding of the nearby area and determine the location, speed, and direction of objects.

**Radar and Lidar**: Radar and Lidar are both wave-based technologies that detect, track, and image the environment. Although these two technologies serve similar purposes, they are different in how they work. These differences then make them appropriate for different scenarios, where one could be favored over the others. Radar uses radio waves to detect and locate objects. Radio waves can have wavelengths from 3 millimeters to thousands of meters. A larger wavelength means a lower frequency and vice versa. Radars that use high frequency, short wave radio waves have a shorter range of detection but yield a much clearer image. Lidar, on the other hand, uses light waves to detect its surrounding objects and track them. Rather than radio waves, Lidar uses light waves to detect its surrounding objects and track them. One of the biggest differences between radar and lidar sensors is the level of accuracy, Lidar being more accurate. Moreover, Radar sensors tend to generate a lot less data as they just return a single point, or a few dozen points. The Lidar sensors sense and transmit lots of data about each individual laser point of range data. Lidar has become increasingly popular in recent years due to its high accuracy compared to other sensing technologies [35]. This superior accuracy creates a clearer map of a vehicle's surrounding area. Yet, it is important to note that Radar has been traditionally widely utilized in various aviation applications, including weather monitoring, and military defense, where it aids in tasks such as collision avoidance, target tracking, and navigation.

Sensing mechanism is often associated solely with hardware devices (called sensors). However, with the advances in sensing technologies, the sensors are increasingly becoming *smart*, that is, a certain level of data-processing abilities are present in the sensing hardware itself. This is referred to as Micro-Electro-Mechanical Systems (MEMS) technology. MEMS technology combines mechanical and electronic components on a small scale, typically at the micro or nanometer level, to create sensors and devices. The following sub-section delineates MEMS further.

*a) Micro-Electro-Mechanical Systems (MEMS) sensors*

It is a technology that incorporates both electronic and moving parts on a microscopic scale. MEMS devices are made up of components between 1 and 100 micrometers in size and generally range in size from 20 micrometers to a millimeter. They usually consist of a central unit that processes data, such as an integrated circuit chip, and several components that interact with the surroundings, such as microsensors [36]. MEMS sensors can sense tiny changes in their environment, be it motion, air pressure, magnetism, or even gases in the air. They relay this information as an electrical signal, making them the sensory organs of the technological world.

MEMS sensors have many applications in the aviation domain [37]. For example, MEMS gyroscopes, accelerometers, and IMUs are used in aircraft and aviation applications, including use in Altitude & Heading Reference Systems (AHRS), standby instrumentation, and flight control surface sensors. MEMS pressure sensors are also widely used in the aviation industry for propulsion/turbomachinery applications, turbulent flow diagnosis, experimental aerodynamics, micro-flow control, and unmanned aerial vehicle (UAV)/micro aerial vehicle (MAV) applications.

MEMS sensors are being applied more and more in Unmanned Aerial Vehicles (UAVs), especially for flight control. They can be used to measure various parameters such as rotation speed, air flow, pressure, force, position, temperature, and vibration [38]. These measurements can be used by the flight control system to make real-time decisions and adjust the aircraft's control surfaces accordingly. For example, MEMS actuators can be used to control leading edge vortex separation and growth, producing a desired aerodynamic force for flight control. MEMS sensors can also be used to detect anomalies in the flight control subsystem and enhance its reliability [39].

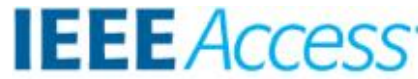

*b) Solid-State Sensors*

Solid-state sensors, often referred to as the sensory cornerstone of modern technology, are devices that employ semiconductor materials to convert a physical property into an electrical signal. They are engineered using advanced semiconductor technology and operate without moving parts, thus enhancing their durability and extending their operational life. These sensors offer several advantages over other types, including high sensitivity, low power consumption, fast response times, and long-term stability. Examples of solid-state sensors include charge-coupled devices (CCD), complementary metal oxide semiconductors (CMOS), and semiconductor lasers.

In the aviation industry, solid-state sensors play a pivotal role across various aviation systems, from monitoring aircraft performance to ensuring flight safety and stability [40]. For example, solid-state temperature sensors are crucial for engine health monitoring, ensuring that critical components operate within safe temperature ranges [41]. Additionally, solid-state pressure sensors contribute significantly to altitude and airspeed calculations, supporting flight navigation and control. These sensors also facilitate real-time data on critical parameters, enabling precise adjustments to the aircraft's control surfaces, engine performance, and navigation systems. Notably, solid-state accelerometers and gyroscopes aid in stabilizing the aircraft, while solid-state pressure sensors help maintain cabin pressure at optimal levels.

(1) FLASH LiDaR (Solid State LiDAR)

Flash LiDAR, also known as Solid State LiDAR, represents an advanced and cutting-edge optical sensing technology within the field of remote sensing and environmental perception. It operates on the principle of light detection and ranging (LiDAR) with a particular focus on the instantaneous illumination of the entire field of view. This innovation leverages solid-state components, such as microelectromechanical systems (MEMS) mirrors and semiconductor lasers, to swiftly capture a three-dimensional spatial profile of the surroundings. Flash LiDAR exhibits a distinctive capacity to generate dense point clouds in a single laser pulse, allowing for real-time, high-resolution mapping and object recognition [42] [43].

*c) Sensor Simulation and Multimodality*

Multi-modal sensor simulation for autonomous aerial mobility is a closely associated research area that aims to develop realistic and scalable methods for testing and validating the performance of sensors and algorithms for UAVs [8]. Sensors such as cameras, lidars, radars, and GPS are essential for enabling autonomous aerial mobility, but they are also subject to noise, interference, occlusion, and other challenges in real-world scenarios [44]. Therefore, it is important to simulate these sensors and their interactions with the environment in a virtual setting, where different conditions and scenarios can be easily controlled and replicated. Multi-modal sensor simulation can also facilitate the integration of different sensor modalities, such as vision and sound, to enhance the robustness and reliability of the proposed autonomy blocks stack. Some of the challenges and opportunities in this field include modeling the physical properties and behaviors of sensors and the environment, generating realistic and diverse synthetic data, evaluating the accuracy and fidelity of sensor simulation, and applying machine learning techniques to improve sensor simulation and data augmentation.

**TABLE 1 Summary of sensed and measured state variables and parameters - for both traditional and AAM based vehicles**

| Sensed and measured states | For traditional aircrafts | For AAM vehicles (e.g. eVTOLs and drones), additional states could include |
|---|---|---|
| Dynamic states | • Airspeed<br>• Altitude<br>• Position<br>• Vertical speed<br>• Heading | • Hover stability<br>• GPS lock status |
| Body axis states | • Angle of attack<br>• Longitudinal acceleration<br>• Lateral acceleration | • Roll rate<br>• Pitch rate<br>• Yaw rate |
| Performance states | • Throttle settings<br>• Fuel flow from engines<br>• Mass | • Battery charge level<br>• Motor RPMs<br>• Power consumption |
| Configuration states | • High lift devices<br>• Landing gear<br>• Speed brake | • Rotor configuration (for tilt-rotor or tilt-wing eVTOLs),<br>• Propeller pitch (for variable-pitch drones),<br>• Payload configuration (for delivery drones or other specialized UAVs). |
| Meteorological (or atmospheric) parameter states | • Wind speed<br>• Wind direction<br>• Static air temperature<br>• Total air temperature<br>• Air pressure | For eVTOLs and drones operating at lower altitudes than traditional aircraft, additional relevant atmospheric parameters could include:<br>• Local weather conditions (rainfall, snowfall)<br>• Microclimate variations (temperature variations near Valley and Slopes, humidity/pressure/temperature changes near water bodies)<br>• Urban heat island effects. |

### 2) Perception

Perception is the higher-level process that comes after sensing. It involves interpreting and making sense of the data collected and preprocessed (to a certain extent) by sensors. Perception algorithms and systems combine data from different sensing streams and analyze it to recognize and identify objects, obstacles, terrain, landmarks, and other relevant features in the environment. Perception also includes estimating the relative positions and velocities of these objects, determining their significance for flight safety and navigation, and predicting their future movements [45]. Assimilating data about vehicle's own states is also part of perception process. In essence, perception enables the autonomous system to understand its surroundings, its relative position and state with respect to those surroundings, and make informed decisions based on that understanding. A non-exhaustive summary of various states variables and parameters that are sensed, measured, and eventually fed to perception blocks is given in Table 1.

#### *a) 3D Object Detection and Tracking*

3D Object Detection is a task in computer vision where the goal is to identify and locate objects in a 3D environment based on their shape, location, and orientation. It involves detecting the presence of objects and determining their location in 3D space in real-time. This task is a crucial first step in the perception component of the autonomy blocks stack of AAM [46].

On the other hand, 3D Object Tracking is a computer vision task dedicated to monitoring and precisely locating objects as they navigate within a three-dimensional environment. It frequently utilizes 3D object detection techniques to pinpoint the objects and establish unique identifications that persist across multiple frames. The goal is to continuously estimate the position and orientation of the object, even in the presence of occlusions, camera motion, and changing lighting conditions.

#### *b) Semantic Scene Understanding*

Unlike 3D object detection which focuses on identifying and locating the objects, semantic scene understanding attempts to analyze objects in the context of the whole scene, unlike object recognition that focuses only on identifying the objects either as 2D or 3D bounding boxes [9]. Semantic scene understanding, therefore, analyzes the objects with respect to the properties like 3D structure of the scene, its layout, and the spatial, functional, and semantic relationships between different objects in the scene [47]. Recent models with high success for scene understanding include [48] [49] [50].

#### *c) Localization*

Localization, in the context of robotics and autonomous aerial systems, refers to the process of determining the precise position and orientation of a vehicle, robot, or object within a given environment. It involves estimating the location relative to a known coordinate system, such as a global map or a local reference frame. Localization is a crucial aspect of navigation and autonomy, as accurate knowledge of the position and orientation is essential for safe and effective operation [10].

Localization can be achieved using various sensors and techniques, often combined for improved accuracy and robustness. Some common sensing technologies used for localization include IMU, LIDAR, and GPS. From methods standpoint, some of the commonly used ones are as follows:

- Visual Odometry (VO): Visual odometry involves using cameras to track visual features and patterns in the environment as the vehicle moves. By analyzing the changes in these features, the system estimates the motion and can update the position and orientation. It's particularly useful in environments with distinctive visual cues.
- Beacon-based Localization: This method involves placing fixed beacons with known positions in the environment. The vehicle uses signals from these beacons to triangulate its position.
- Radio Frequency (RF) Localization: Using radio signals, such as Wi-Fi, Bluetooth, or RFID, vehicles can estimate their positions based on signal strength and the known locations of access points or transmitters.
- Particle Filters and Kalman Filters: These are probabilistic filtering techniques that combine measurements from different sensors to estimate the vehicle's position and orientation while accounting for uncertainty and sensor noise.

Accurate localization is essential for autonomous vehicles, drones, robots, and other systems to operate safely and effectively. By knowing their precise position, these systems can plan routes, avoid obstacles, and make informed decisions during their tasks.

##### (1) Visual Odometry

Visual odometry (VO) is the process of incrementally estimating the pose of the vehicle/robot by examining the changes that motion induces on the images of its onboard cameras [51]. This technique estimates the motion and position of a vehicle by using only the images captured by a camera attached to it. VO is useful for navigation and obstacle avoidance in various environments, especially where other sensors or systems are not available or reliable, such as indoors, underwater, or in space. VO can also provide 3D vision and rich information about the surroundings.

VO can be classified into different types based on the type of camera used, such as stereo, monocular, omnidirectional, or RGB-D cameras. Each type has its own advantages and disadvantages in terms of cost, accuracy, calibration, and scale estimation. VO can also be approached in different ways based on the method of image analysis, such as feature-based, appearance-based, or hybrid methods. Feature-based methods extract and match distinctive features between image frames, such as corners, lines, or curves. Appearance-based methods use pixel intensity values to measure the changes in the image appearance. Hybrid methods combine both feature and appearance information to improve the robustness and efficiency of VO.

VO, however, also faces many challenges that affect its performance and reliability. Some of these challenges are related to the computational cost of image processing, the lighting and imaging conditions of the environment, the presence of noise and blurs in the images, the lack of texture or features in the scene, and the drift accumulation over time. Therefore, VO often requires integration with other sensors or systems, such as GPS, INS, or laser sensors, to enhance its accuracy and stability.

(2) SLAM

Simultaneous Localization and Mapping (SLAM) is a technique that combines the process of building a map of the environment with estimating the vehicle's position within that map [52]. It's commonly used in robotics to navigate in unknown or changing environments. It refers to a comprehensive approach used in robotics and autonomous systems to simultaneously create a map of an unknown environment while estimating the position and orientation of the vehicle or sensor within that environment [53].

As a method, SLAM represents the general concept of addressing the challenge of mapping an unknown area while navigating within it. It involves the integration of sensor data, such as LIDAR scans, camera images, and IMU measurements, to build a coherent map of the environment. As a technique, SLAM involves a specific set of algorithms and computational processes that combine sensor measurements, motion models, and probabilistic methods to iteratively update the map and the estimated position as the vehicle moves through the environment. These algorithms handle uncertainties, noise, and errors in sensor data to maintain accurate localization and mapping over time.

In summary, SLAM is a concept or method that encompasses the overarching idea of mapping and localization simultaneously. It is also a specific technique involving algorithms and computational strategies to achieve that goal in practice.

SLAM vs VO: The main difference between VO and SLAM is that VO mainly focuses on local consistency and aims to incrementally estimate the path of the camera/robot pose after pose, and possibly performing local optimization [54]. On the other hand, SLAM aims to obtain a globally consistent estimate of the camera/robot trajectory and map. In other words, VO is concerned with estimating the motion of the camera/robot in real-time, while SLAM is concerned with building a map of the environment while keeping track of the camera/robot's location within it. Both techniques have their own strengths and weaknesses and are often used together to achieve accurate and robust navigation and localization.

*d) Mapping*

Mapping, in the context of robotics and autonomous systems, refers to the process of creating a representation or model of the environment. The goal of mapping is to capture spatial information about the surroundings, including the locations of objects, obstacles, landmarks, and other relevant features [11]. Mapping is a crucial aspect of navigation, exploration, and understanding for autonomous systems. By having an accurate and up-to-date map of the environment, autonomous aerial vehicles can make informed decisions, plan optimal paths, avoid obstacles, and navigate effectively. Mapping can be performed in various domains, such as indoor environments, outdoor landscapes, or even underwater areas.

There are different types of mapping techniques and technologies, each suited for specific environments and applications:

- **Occupancy Grid Mapping**: This method divides the environment into a grid of cells and assigns probabilities to each cell based on the likelihood of occupancy. It's commonly used for representing indoor environments and detecting obstacles [55].
- **Feature-Based Mapping**: This approach focuses on identifying and mapping specific features or landmarks in the environment, such as corners, edges, or distinctive objects [56]. Feature-based maps can be useful for navigation and localization.
- **Topological Mapping**: Instead of representing the environment in a geometric way, topological mapping focuses on capturing the connectivity and relationships between different locations or areas [57]. It's often used for high-level navigation planning.
- **3D Mapping**: This involves creating a three-dimensional representation of the environment using technologies like LIDAR or depth-sensing cameras [58]. 3D maps provide more detailed information about the environment's structure.
- **Semantic Mapping**: In addition to geometry, semantic mapping includes information about the types and categories of objects in the environment [59]. This means that the maps contain semantically meaningful objects, which can provide actionable information for various applications [60]. This is particularly useful in real-world applications where understanding the environment goes beyond just knowing the geometry of the surroundings. This can help in understanding the context and making more informed decisions.
- **SLAM**: SLAM, also covered under Localization subsection above, combines mapping and localization. It involves building a map of the environment while simultaneously estimating the robot's or vehicle's position within that map [52].

Mapping can be done in real time as a robot or vehicle moves through the environment (online mapping) or offline by processing recorded sensor data (offline mapping). Regardless of the approach, mapping plays a crucial role in enabling autonomous systems to interact with their surroundings in a meaningful and intelligent way.

- **Online Mapping**: Online mapping is a real-time process in which a robotic system, equipped with various sensors such as LIDAR, cameras, and IMUs, creates a representation of its environment as it navigates. This process involves concurrently estimating the system's position and orientation while updating the map. The robot employs algorithms like SLAM (Simultaneous Localization and Mapping) to fuse sensor measurements, motion models, and probabilistic methods to iteratively build a coherent map of the environment while maintaining an accurate estimate of its own location. Online mapping is essential for tasks requiring live interaction with the environment, enabling the robot to adapt to dynamic changes and navigate in real time.

- **Offline Mapping**: Offline mapping, on the other hand, involves the post-processing of recorded sensor data to generate a map of an environment after the robotic system has completed its exploration or mission. The raw sensor data, such as LIDAR scans and camera images, are collected during the robot's operation and then processed offline using mapping algorithms. These algorithms analyze the accumulated data, align sensor measurements, and reconstruct the environment's features and geometry. Offline mapping is advantageous for situations where real-time constraints are less critical and where a more accurate and refined map can be generated through careful data processing and optimization, without the pressures of immediate decision-making.

Both online and offline mapping methods have their own advantages and trade-offs, and their suitability depends on the specific application, computational resources, and the required level of accuracy and responsiveness. Online mapping is suitable for scenarios where real-time adaptation and navigation are critical, while offline mapping offers the opportunity to refine and analyze collected data to create high-quality maps for subsequent analysis or planning.

#### 3) Planning

"Path Planning" and "Mission Planning" are two distinct concepts in the context of autonomous aerial mobility or robotics in general. While they are related and often work together, they serve different purposes:

##### *a) Path Planning*

Path planning refers to the process of determining an optimal path or trajectory for a vehicle (e.g., drone or other UAV) to navigate from its current position to a specific goal or destination while avoiding obstacles and adhering to certain constraints. The path planning algorithm takes into account the vehicle's dynamics, environment information (obstacles, terrain, etc.), and other relevant factors to calculate the most efficient and collision-free path. The goal of path planning is to find a feasible and safe trajectory that guides the vehicle from start to end while optimizing for criteria like time, energy consumption, or smoothness. Some of the commonly utilized path planning algorithms for aerial vehicles include:

- Dijkstra and A* algorithms: These are the most commonly used methods in autonomous mobile robots. While the Dijkstra algorithm determines the shortest path between two nodes, the A* algorithm also finds the shortest path by using heuristic approaches [61].
- Artificial Potential Field methods: These are conventional global path planning algorithms [62].
- Ant Colony Algorithms: These are optimization algorithms inspired by the behavior of ants. They have been used in drone path planning [62] [63].
- Rapidly Exploring Random Trees (RRT): This is a data structure and method that is designed for efficiently searching nonconvex, high-dimensional spaces [63].
- Swarm Optimization Algorithms: These include ant colony optimization (ACO), fruit fly optimization algorithm (FOA), artificial bee colony (ABC), and particle swarm optimization (PSO). They provide intelligent modeling for drone path planning [63].
- Genetic Algorithm: This is a search heuristic that is inspired by Charles Darwin's theory of natural evolution [64].
- Deep Neural Networks: These are artificial neural networks with multiple layers between the input and output layers [65].
- Hybrid Algorithms: These combine two or more algorithms to get better results.

Each of these algorithms has its own advantages and is used based on the specific requirements of the application scenario in which the path planning is to be conducted by the UAVs.

##### *b) Mission Planning*

Mission planning, on the other hand, is a higher-level concept that involves defining and organizing a set of tasks or objectives that an autonomous system needs to accomplish to achieve a specific goal. It involves determining the sequence of actions, waypoints, and goals that the vehicle or robot should follow to complete its mission successfully. Mission planning encompasses multiple aspects, including path planning for individual segments, task allocation, resource management, and coordination among multiple vehicles or agents if applicable. It considers the overall mission objectives and optimizes the allocation of resources and tasks to achieve the mission's end goal efficiently. Fig 4 below shows the various factors that Mission Planning takes into account. Path Planning, therefore, can be considered a sub-module within the Mission Planning process.

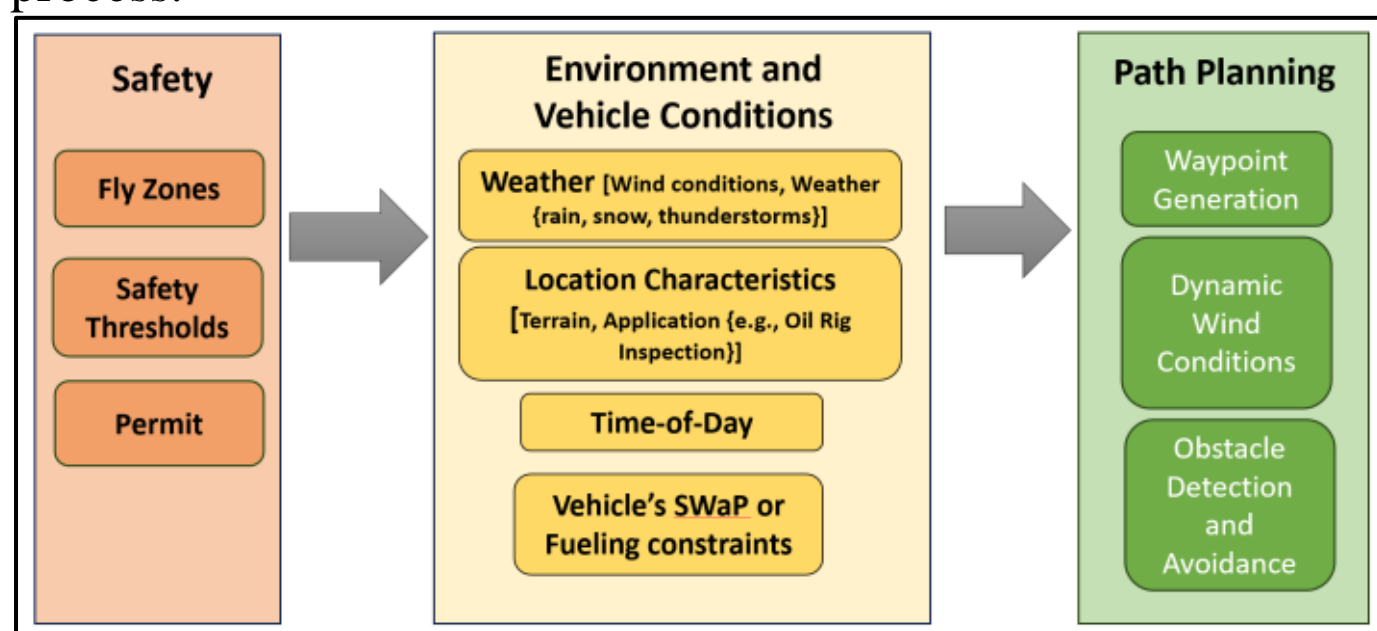

**FIGURE 4** **Three stage process of Mission Planning**

In summary, path planning deals with finding the best trajectory or path for a single vehicle to navigate through its environment, while mission planning involves defining a higher-level strategy that organizes multiple tasks and objectives to achieve a specific mission or goal. Path planning is a component of mission planning, as a successful mission often requires the autonomous vehicle to navigate through various paths and trajectories to accomplish its tasks along the way.

#### 4) Controls

The fourth and the last functional block of an autonomous system framework is actuation. Actuation translates to controls in case of aerial autonomy. After the environment is perceived and understood, the vehicle needs to maneuver accordingly. This is achieved through control algorithms. These algorithms determine how the vehicle should move based on the dynamically perceived environment, vehicle's aerodynamics in the meteorological conditions at the moment.

In case of UAV like a drone, the primary controls include:

- Roll: This allows the drone to move to the right or the left along the roll axis that runs from the front of the drone to the back of the drone.
- Pitch: This tilts the drone forward or backward.
- Yaw: This rotates the drone clockwise or counterclockwise, allowing it to make circles or patterns in the air.
- Throttle: This controls the amount of power sent by the battery to the motors, which makes the drone go faster or slower.

Similarly, in case of a passenger carrying commercial aircraft, primary flight controls are:

- Ailerons: These control the rolling motion of the aircraft through the longitudinal axis.
- Elevator: This controls the pitch of the aircraft through the lateral axis.
- Rudder: This controls the yaw of the aircraft through the vertical axis.

And the secondary flight controls:

- Flaps and Slats: These help to slow down the aircraft for landing and help to reduce the ground roll on take-off.
- Trim Control Surfaces: These reduce the effort the pilot has to apply to fly the aircraft.
- Spoilers and Speed Brakes: These assist the pilot in roll and speed and lift reduction.

In a flight control system, the mixer and actuators play crucial roles. The mixer takes force commands (e.g., turn right) and translates them into actuator commands which control motors or servos. For example, in a plane with one servo per aileron, this means to command one of them high and the other low. Depending on the complexity of the aircraft or UAV, the cyclic and collective may be linked together by a mixing unit, a mechanical or hydraulic device that combines the inputs from both and then sends along the "mixed" input to the control surfaces to achieve the desired result. Actuators are devices that convert energy into motion. In an aircraft flight control system, actuators convert hydraulic pressure or electrical signals into control surface movements. For instance, in power-by-wire systems, electrical actuators are used in favor of hydraulic pistons. The power is carried to the actuators by electrical cables. These components work together to ensure precise control of the aircraft's movement and behavior.

Control systems in traditional aircraft are generally divided into two categories, open- and closed-loop systems. A common type of controller used in these systems is the Proportional-Integral-Derivative (PID) controller, which is a closed-loop control system. The PID controller adjusts the control inputs to the aircraft based on the error between the desired and actual states of the aircraft. In addition to PID, other control strategies like Linear Quadratic Regulator (LQR), and neural networks have also been experimented [66]. In the case of drones, the control algorithms determine the rotational speed of the propellers that guide the drone to a particular position in 3D space. Even more fundamentally, the asymptotic stability of the UAV in the air is ensured by "kernel control law". Several algorithms have been analyzed for autonomous quadrotors including their advantages and disadvantages: PID, LQR, Sliding mode, Backstepping, Feedback linearization, Adaptive, Robust, Optimal, L1, $H\infty$, Fuzzy logic and Artificial neutral networks [67]. Fundamentally, the objective of the controller is to reduce the error between the estimated and desired states – desired states being fed to the controller in the form of reference commands.

In the case of autonomous operations, the higher-level control commands are guided by the path planning layer – that generates waypoints to follow specific trajectories. Path planning layer also encompasses battery management system to optimize flight time and return to the base or charging station when required. All these four functional blocks of autonomously navigating aerial vehicle work together to dynamically detect obstacles and avoid collision by changing the trajectory in response to perceived obstacle [68]. This advanced capability closes the fully autonomous operations loop of an aerial vehicle [69]. This capability is often referred to as *Obstacle Detection and Avoidance*.

### B. ORCHESTRATION OF AUTONOMY BLOCKS TECHNOLOGY WITH THE ECO-SYSTEM

Autonomy blocks, the core components of autonomous aerial vehicles, are intricately designed and meticulously integrated into a complex ecosystem. This intricate fashion encompasses a delicate interplay between the hardware platform, cloud platform, on-board computing systems, and an underlying operating system that runs on the on-board compute that underpin them.

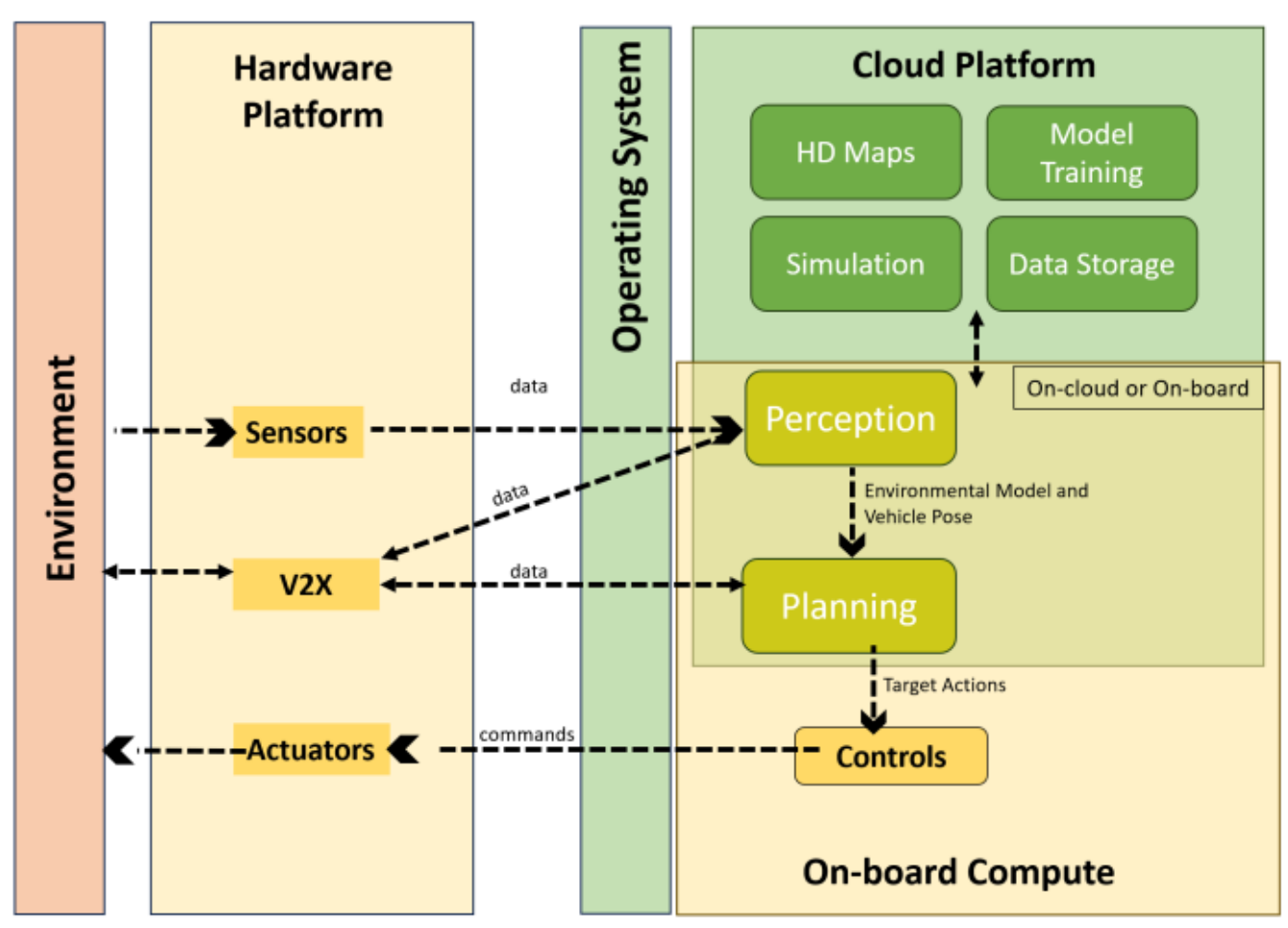

**FIGURE 5 Orchestration of Hardware, Software, Cloud and On-board compute platforms where Autonomy Block technology operates.**

These autonomous AAM vehicles are engineered to seamlessly support a myriad of protocols and standards, enabling the vehicle to interact with a diverse array of sensing devices and establish efficient communication with other entities, such as fellow vehicles, ground stations, vertiports, and charging infrastructure.

This multifaceted orchestration is essential to ensure the smooth and safe operation of autonomous vehicles, highlighting the depth of engineering and innovation involved in their development. Fig 5 depicts the interplay of these components. Note that the Perception and Planning blocks for a particular vehicle type can either be housed in the cloud platform or on the on-board computer depending on the size, weight, and power (SwaP) constraints of a specific UAV design and the specific application scenario the vehicle is designed for and operated in. In summary, this orchestration (also called flight stack or autopilot) serves the function of acquiring data from sensors, regulating motor functions to maintain UAV stability, and facilitating communication for ground control and mission planning.

### C. AUTONOMY-ASSISTED PHASES OF FLIGHT

The Autonomy Blocks, as outlined in our proposed framework, are modular and can be utilized in one or more flight phases. Although there is a constant requirement for sensing mechanisms to operate throughout all flight phases, certain Autonomy Blocks may not be necessary in specific phases. For instance, in a commercial aircraft, once the cruising phase has begun, trajectory planning is seldom required. However, for firefighting survey drones, the path planning task is dynamic and thus the block is engaged for the entire duration of the flight. At its core, different phases of flight are marked by the changes in speed and direction of the vehicle, which in turn translates to acceleration. At its core, there are three types of acceleration that help maneuver an aircraft or a drone: 1) Linear Acceleration, 2) Radial Acceleration, and 3) Angular Acceleration.

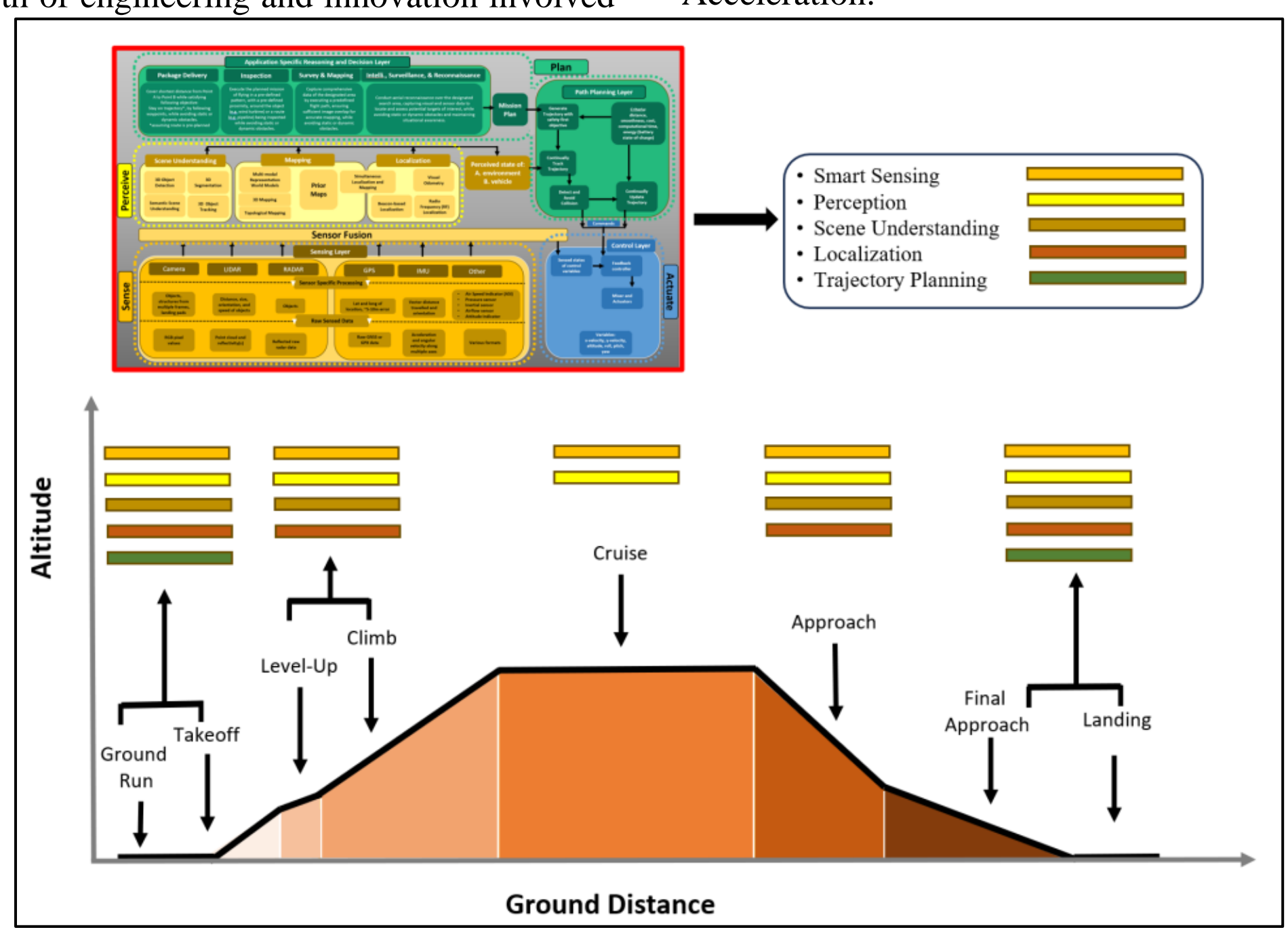

**FIGURE 6 Mapping of various autonomy blocks to the phases-of-flight**

- Linear Acceleration: This is observed when there's a change in speed along a straight line. In the context of a quadrotor drone, this happens during actions like take-off, landing, or when there's an adjustment in the throttle setting. It's determined by the sum of gravity, thrust from the motors, and linear friction force causing drag.
- Radial Acceleration: This is associated with a shift in direction. For example, when the drone performs a sharp

turn or rapidly changes altitude. In quadrotor dynamics, this can be linked to alterations in the flight path due to external influences or navigational instructions.

- Angular Acceleration: This is the result of a change in both speed and direction at the same time. In quadrotor control systems, angular acceleration is vital during complex maneuvers like spins and climbing turns and for maintaining stability and control during flight.

These accelerations are crucial for accurately tracking aggressive quadcopter trajectories and are utilized to refine control algorithms for robust tracking of linear and angular accelerations.

Fig 6 illustrates the engagement or utilization of various autonomy blocks during different flight phases for an autonomously operating commercial aircraft, as an example. It's important to note that the usage pattern of the proposed modular autonomy blocks varies from one type of aerial vehicle to another. This variation can also depend on the specific application for which the aerial vehicle is used. For instance, a drone conducting a site inspection in a GPS-denied environment [70] would require robust on-board localization and trajectory planning capabilities. Conversely, a drone tasked with delivering a meal in urban areas, where the starting and ending points are precisely mapped out, may not require functionality for localization in GPS-denied environments. Furthermore, depending on the types of environments and the types of applications an aerial vehicle is employed for, the on-board compute capability – which is required to house one or more autonomy blocks – vary.

In the following subsection, we further delineate the acceleration characteristics in different phases of flight and intentionally offer two contrasting examples for each – one from traditional aviation scenario and another from the fast-growing UAV sector of aerial mobility (e.g. eVTOLS and drones).

### 1) Take-off and Climb Phase

During take-off, the airplane accelerates from zero ground speed to a speed at which it can lift itself from the ground. The thrust must exceed drag for acceleration to take place. Once the aircraft has lifted off and begins to climb, some of the excess thrust goes into climbing, so horizontal acceleration decreases. Both vertical and horizontal accelerations are significant during the take-off and climb phase of an aircraft's flight.

The process of take-off and climb for an eVTOL (Electric Vertical Take-off and Landing) aircraft differs slightly from that of a traditional aircraft, given its capability to ascend vertically, similar to a helicopter. In the take-off phase, the eVTOL moves from a standstill to a speed that allows it to rise vertically off the ground. This is accomplished by amplifying the thrust of its electric motors, which drive its propellers or rotors. The eVTOL must generate thrust that surpasses its weight to lift off. Following lift-off, the eVTOL enters the climb phase. During this stage, the eVTOL transitions from hovering to forward flight. This transition involves increasing thrust and adjusting the propellers' or rotors' angle to provide both lift and forward thrust. As with conventional aircraft, both vertical and horizontal accelerations play a significant role during the take-off and climb phase of an eVTOL's flight.

### 2) Cruise / Mission Phase

In the cruising phase of an aircraft's flight, the plane maintains a steady airspeed and altitude. The vertical forces, namely weight and lift, are in equilibrium, leading to nearly zero vertical acceleration or velocity. This is because the aircraft isn't ascending or descending during this phase. The main forces in play during cruise are thrust and drag, which are balanced in the horizontal axis.

While an aircraft in its cruise phase maintains a steady airspeed and altitude with balanced forces of thrust and drag, a drone inspecting a transmission line operates quite differently. It hovers and maneuvers around the lines, collects detailed data, handles interference from the lines, all while ensuring safety. The autonomy blocks in action during such mission execution are different from the ones that would be employed for cruising phase of a commercial jet.

### 3) Final Approach and Landing Phase

In the final approach phase, the aircraft follows a descent path towards the runway. There might be a slight increase in thrust to make minor speed corrections. However, this stage of flight should not necessitate substantial thrust increases. As the aircraft lands, it makes contact with the runway at landing speed and slows down to a standstill. The deceleration is achieved through braking, aerodynamic drag, ground friction, and potentially reverse thrust, bringing the plane's speed to a halt. This represents a state of decelerated motion. Therefore, during the final approach and landing phases of an aircraft's flight, there is significant deceleration and minimal acceleration. All of the autonomy blocks modules need to be engaged during this phase of flight.

This process varies in the case of other types of aerial vehicles. For example, during the final approach phase, the eVTOL shifts from forward flight to a hoover in preparation for landing. This shift involves a decrease in forward speed while the eVTOL descends towards its landing destination. The eVTOL's electric propulsion system is used to manage its descent and ensure stability. As the eVTOL approaches the ground, it might slightly increase its thrust to counteract the ground effect (a phenomenon that can cause the aircraft to 'float' when near the ground), guaranteeing a smooth and controlled landing. When the eVTOL touches down, it decelerates until it has come to a complete stop. This deceleration is achieved through a mix of reduced thrust and aerodynamic drag. Unlike conventional aircraft, eVTOLs typically lack reverse thrust capabilities or mechanical brakes, so they depend on their motors and propellers for deceleration.

### 4) Autonomous Taxi, Take-Off and Landing (ATTOL)

One example of ATTOL is a project (with the same name) conducted by Airbus in the year 2020. The Autonomous Taxi, Take-Off and Landing (ATTOL) project by Airbus has been successfully completed after an extensive two-year flight test program. This achievement marks a significant milestone in the traditional aviation industry as it demonstrates the capability of a commercial aircraft to taxi, take-off, and land autonomously using cutting-edge on-board image recognition technology [71]. The project involved more than 500 test flights, with approximately 450 flights aimed at collecting raw video data to refine the underlying algorithms. A set of six distinct test flights, each consisting of five take-offs and landings, were specifically designed to evaluate the autonomous flight capabilities. This successful demonstration of autonomous capabilities signifies not only the industry's commitment but also the rapidly approaching future of autonomous aviation.

## V. APPLICATIONS OF UAVs

Autonomy blocks framework presented in this article can be conceptually mapped with the phases-of-flight and applications as shown in Fig 7. In other words, the modular design of autonomy blocks enables the mapping of these three dimensions (block type, application type, and phase of flight) in a many-to-many fashion. As an example, a "Delivery" drone, in its descend phase, can have multiple autonomy blocks (sensing, perception, obstacle avoidance) engaged in enabling its navigation to the pre-planned destination. Also, one or more autonomy blocks can be utilized for an application's specific phase-of-flight.

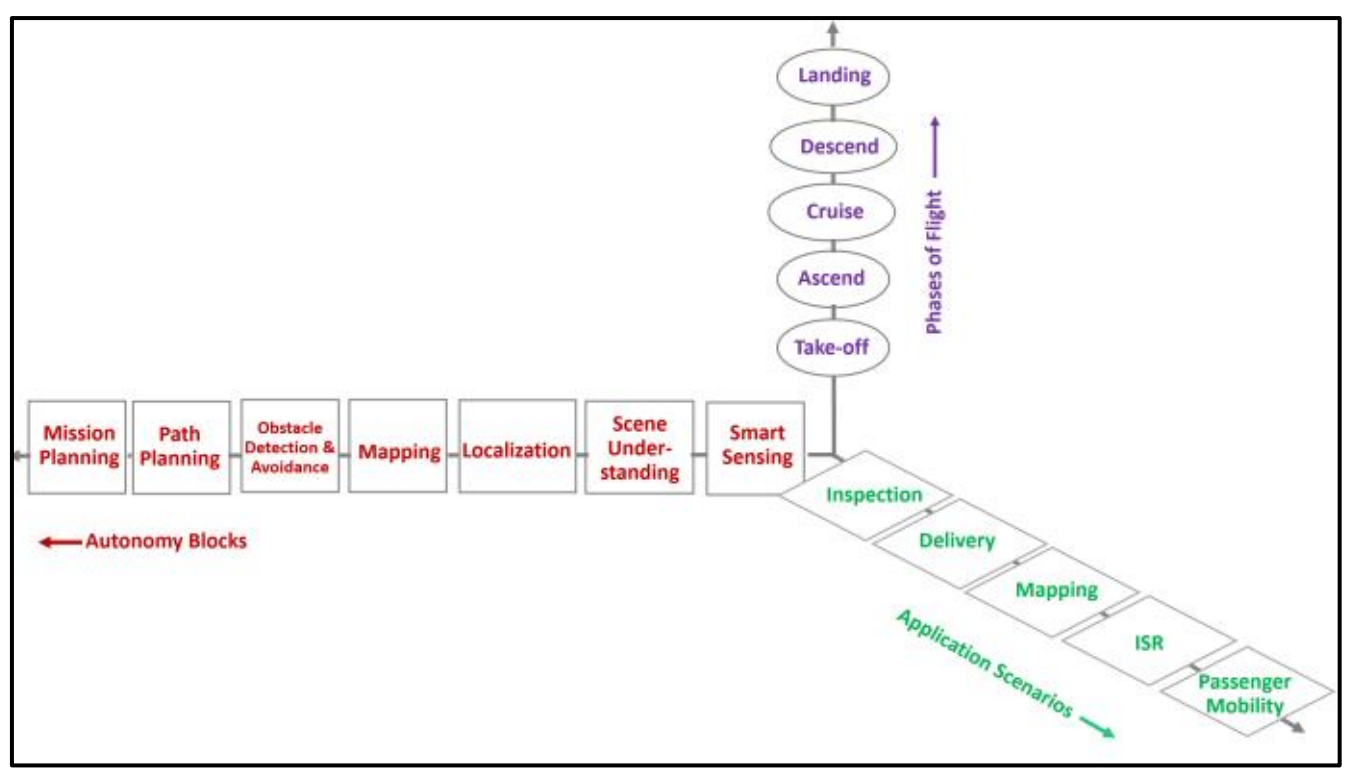

**FIGURE 7 Intersection of three spheres where technology meets demand: Autonomy Blocks, Application Scenarios, and Phases-of-Flight.**

UAVs can be categorized based on various attributes including lift technology (vectored thrust, multirotor, lift plus cruise), propulsion type (fully electric, hybrid, electric hydrogen), mode of operations (autonomous, piloted), and applications. This section details the applications categorization. UAVs of different shapes and sizes have a plethora of applications across various domains and sectors. In this article, we provide three different ways of categorizing the applications of UAVs:

1. **Domains**: this classification is specific to the purpose for which the UAV is designed.
2. **Sectors**: this classification is specific to the sector that the UAV serves.
3. **Scenarios**: one or more UAVs of different types and designs can be employed in serving a single scenario. We demonstrate this using three specific scenarios.

Flowchart in Fig 8 enlists the applications within the above three types of categorizations.

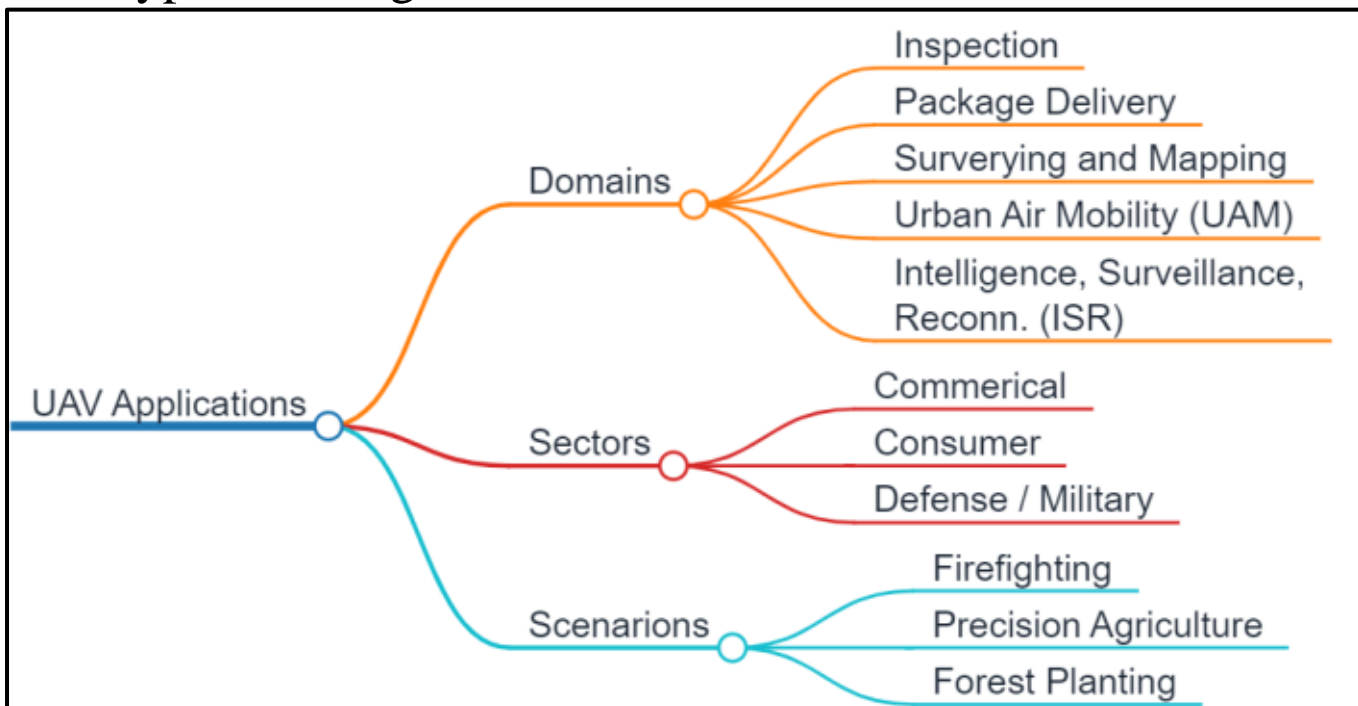

**FIGURE 8 UAV Applications, Categorized.**

### A. DOMAINS

The five application domains, in which the fast growing UAVs sector can be classified, are: Passenger Mobility (UAM), Inspection, Package Delivery, Surveying and Mapping, and Intelligence, Surveillance, Reconnaissance (ISR). The following subsections discuss these in more details:

#### 1) Urban Air Mobility (UAM)

Urban Air Mobility (UAM) represents a system for the air transportation of passengers and cargo within a metropolitan area, including operations over densely populated urban areas [72]. It also includes an urban air ambulance service which is a part of UAM that prioritizes the use of air transportation for emergency medical services. The FAA perceives the burgeoning cities and increasing population density as an opportunity to revolutionize air transportation by materializing UAM [73]. Similarly, government and private sector around the world are considering urban air transportation a worthwhile effort to manage the growing mobility requirements of the increasing urban population and alleviate congestion of road transportation networks. The UAM ecosystem expands transportation networks to incorporate both crewed and uncrewed aircraft and explores solutions utilizing agile infrastructure and diverse operations. The UAM Program addresses interactions with existing Air Traffic Control (ATC) and the role of cooperative traffic management concepts explored in Unmanned Aircraft System (UAS) Traffic Management (UTM), further details in Section VI.A below.

The concept development considers the introduction of new aircraft types, such as electric Vertical Takeoff and Landing (eVTOL) [74] [75], with an increasing level of autonomy and the data exchanges they require. The majority of eVTOL aircraft currently in design or prototype stages utilize electric or hybrid electric propulsion systems. These systems comprise Energy Storage Systems (ESS), typically large Lithium-Ion battery modules, and associated Battery Management Systems (BMS) connected to various electric motors and propellers.

Through collaboration with NASA [76] and industry partners, the UAM Program identifies and validates critical paths to determine minimally viable operations for the near future. Engineering and analysis for the UAM Program focus on the unique traffic management requirements, procedures, airspace design, and policies for the operational environment.

### 2) Intelligence, Surveillance, Reconnaissance (ISR)

ISR is a military and security concept that involves gathering, analyzing, and utilizing information to support decision-making and operations. ISR activities aim to provide a comprehensive understanding of the operational environment, enemy activities, potential threats, and other relevant factors. Here's a breakdown of each component of ISR:

- Intelligence: Intelligence refers to the collection, analysis, and interpretation of information to gain insights into the intentions, capabilities, and activities of potential adversaries or entities of interest. This information can come from various sources, such as signals intelligence (SIGINT), human intelligence (HUMINT), open-source intelligence (OSINT), and more. Intelligence helps decision-makers understand the context in which they operate, identify potential risks, and develop effective strategies.
- Surveillance: Surveillance involves the continuous observation and monitoring of areas, assets, or individuals to gather real-time information. This can be done using various technologies, including cameras, sensors, drones, satellites, and aircraft. Surveillance provides live updates on activities, movements, and changes in the environment, allowing for rapid response and decision-making.
- Reconnaissance: Reconnaissance, often abbreviated as "recon", is the act of exploring and gathering information about an area, route, or target. This can involve sending personnel, vehicles, aircraft, or unmanned systems into an area to assess its characteristics, potential threats, and opportunities. Reconnaissance helps to identify potential targets, vulnerabilities, and valuable information that can inform future actions.

ISR activities are critical in military, security, and law enforcement operations for a variety of purposes:

- **Search and Rescue**: ISR capabilities can assist in locating and providing assistance to individuals in distress, whether in a combat zone or a disaster-stricken area.
- **Strategic Planning**: Intelligence gathered through ISR supports long-term strategic planning, including assessing the capabilities and intentions of potential adversaries.
- **Force Protection**: ISR helps protect personnel and assets by identifying potential threats and allowing for preemptive action or evacuation.
- **Situational Awareness**: ISR provides a real-time understanding of the operational environment, enabling decision-makers to respond to emerging threats or changes effectively.
- **Target Identification and Tracking**: ISR supports the identification and tracking of potential targets, such as enemy forces, vehicles, or infrastructure, which is crucial for planning and executing operations.
- **Battlefield Management**: In military contexts, ISR aids in managing the battlefield by providing commanders with timely information to allocate resources and adjust strategies.

ISR plays a pivotal role in enhancing decision-making, operational effectiveness, and safety across a wide range of security and defense scenarios. UASs and UAVs play an increasingly crucial role in ISR operations [77] [78].

### 3) Inspection, Package Delivery, Surveying and Mapping

Drones offer versatile, cost-effective, and efficient solutions in all these three domains. Various applications with these three domains are summarized in the table below.

**TABLE 2 Application with inspection, package delivery, surveying and mapping domains**

| Domain | Inspection | Package Delivery | Surveying and Mapping |
|---|---|---|---|
| **Applications within Domain** | Drones enable high-quality data capture in various applications, including:<br>• Wildfire monitoring<br>• Oil and Gas pipelines inspection<br>• Oil Extraction sites (in the ocean) inspection<br>• Electricity transmission networks – inspect vegetation and transmission components' health.<br>• Wind turbine inspection | Drones mobilize goods – with lesser energy and time consumption – in a wide range of applications, including:<br>• Critical Medicines delivery<br>• Consumer Package deliveries<br>• Forest Planting Support<br>• Maritime communication: facilitate the transmission of data from a ground station to a mobile vessel even when a Line of Sight (LoS) path is unavailable. | Drones capture high-resolution aerial imagery, 3D models, and geospatial data for various applications, including:<br>• Land surveying,<br>• Terrain mapping,<br>• Environmental monitoring,<br>• Urban planning and construction<br>• Weather forecasting<br>• Precision agriculture<br>• Real-time road traffic monitoring<br>• Planetary exploration in the space (surface and atmosphere) |

### *B. SECTORS*

#### 1) Commercial

- Agriculture and Surveying Drones: These drones are equipped with Level 3 and above autonomous flight capabilities for precision agriculture and surveying.
  - DJI Agras Series: Drones designed for precision agriculture and crop spraying.
  - SenseFly eBee Series: Fixed-wing drones for mapping and surveying large areas.
- Delivery Drones: These delivery drones are designed to navigate autonomously to deliver packages to specific locations.
  - Amazon Prime Air: Drones designed for package delivery.
  - Wing by Google: Delivery drones for transporting goods to consumers.
- Inspection Drones: These drones often come with autonomous flight features for conducting inspections of infrastructure and confined spaces.
  - DJI Matrice Series: Professional drones equipped with sensors for infrastructure inspection.
  - Flyability Elios: Collision-tolerant drones for confined space inspection.
- Mapping and 3D Modeling Drones: These drones can execute pre-planned autonomous flights for mapping and creating 3D models. These drones are equipped with cameras and sensors to create high-resolution maps and 3D models of the terrain.
  - DJI Phantom Series: Used for aerial mapping, creating 3D models, and topographical surveys.
  - Parrot Anafi USA: UAVs for mapping, inspection, and situational awareness.

#### 2) Defense

- Reconnaissance and Surveillance UAVs: These UAVs are capable of autonomous flight for reconnaissance and surveillance missions. They can also be armed for strikes, but human operators often make the final decision to engage targets. They can carry sensors, target designators, offensive ordnance, or electronic transmitters designed to interfere with or destroy enemy targets.
  - Predator/Reaper: Long-endurance UAVs for intelligence gathering, surveillance, and reconnaissance with strike capabilities.
  - Global Hawk: High-altitude, long-range UAVs for wide-area surveillance and data collection.
  - Shadow: Tactical UAVs used for real-time reconnaissance, target tracking, and battle damage assessment.
- Combat UAVs (UCAVs – Unmanned Combat Aerial Vehicles): Similar to the Predator/Reaper, they are semi-autonomous for strike missions, but human operators typically retain control over engagement decisions.
  - MQ-9 Reaper: Multi-role UAV with strike capabilities for offensive operations.
  - X-47B: Experimental UAV for carrier-based operations, including strike and reconnaissance.
- Drone Swarms: Drone swarms can exhibit different levels of autonomy, from individual drones following a predetermined path to more advanced collaborative behaviors.
  - Swarms of small UAVs for collaborative missions, surveillance, and coordinated attacks.
  - Perdix: Micro-drones used in swarm formations for various military applications.

#### 3) Consumer

- Aerial Photography (Selfie) and Videography Drones: these consumer and prosumer drones offer various levels of autonomy, such as follow-me modes, waypoint navigation, and obstacle avoidance.
  - DJI Phantom Series: Consumer-level drones equipped with cameras for photography and videography.
  - DJI Mavic Series: Foldable drones with advanced camera systems for professionals.
- Racing and Acrobatic Drones: These drones are typically flown manually by experienced pilots and may not focus as much on autonomous features.
  - FPV racing drones: Customizable drones for competitive racing and aerial maneuvers.
  - Betaflight HX100: Micro-sized drones for indoor and outdoor acrobatics.
- Toy Drones: These drones are generally more basic and may not include advanced autonomous features.
  - Hubsan X4: Small drones designed for fun and entertainment.
  - Ryze Tello: Programmable drones suitable for educational purposes.

### *C. SCENARIOS – DEEP DIVE EXAMPLES*

The following three sub-sections describe multitude of use cases of drones for three specific applications – Fire Fighting, Precision Agriculture, and Forest Planting. The rationale behind categorizing the applications of UAVs in 'scenarios' categories is to demonstrate the versatility of applications that UAVs offer within a single use-case (e.g. fire-fighting).

#### 1) Fire fighting

UAVs, specifically drones, play a crucial role in firefighting scenarios by providing valuable data, real-time monitoring, and support to firefighting teams. Drones enhance situational awareness, improve decision-making, and aid in managing firefighting operations in both urban and wildland environments. Drones contribute to firefighting in following ways:

- **Aerial Surveillance and Situational Awareness** (Rapid Response): Drones provide aerial views of the fire scene, allowing firefighters to assess the size, spread, and intensity of the fire. Real-time imagery and video feed enable commanders to make informed decisions about resource allocation, evacuation routes, and deployment strategies.

- **Early Detection and Monitoring – Thermal imaging**: Drones equipped with thermal cameras can detect hotspots and areas of intense heat, even in smoke-filled environments. This early detection helps prevent the fire from spreading and allows firefighters to focus on containment efforts.
- **Real-Time Mapping and Assessment**: Drones can create detailed live maps of the fire-affected area, helping firefighters understand the terrain, identify safe zones, and plan evacuation routes, thereby enhancing situational awareness. These maps can also assist in assessing the fire's progress and estimating its potential impact.
- **Search and Rescue Operations**: Drones aid in locating missing individuals in areas affected by fires. Drones can navigate hazardous environments to locate people in distress. Thermal cameras can detect human heat signatures, helping rescue teams find and save people in danger.
- **Safety Monitoring**: Drones allow firefighters to remotely monitor fire behavior and conditions in hazardous areas. This reduces the risk to personnel and provides critical information for making timely evacuation decisions.
- **Communication Support**: Drones equipped with communication equipment can establish temporary communication networks in areas where traditional infrastructure has been compromised, enabling better coordination among firefighting teams.
- **Smoke and Air Quality Monitoring**: Drones can measure air quality and smoke concentration levels, helping authorities provide accurate information to residents and manage potential health risks.
- **Aerial Suppression – Water and Retardant Drops**: Specialized drones equipped with firefighting payloads, such as water or fire retardant, can assist in suppressing flames and creating firebreaks in hard-to-reach areas.
- **Post-Fire Assessment**: After the fire is extinguished, drones can conduct post-fire assessments to evaluate the extent of damage, assess structural integrity, and aid in recovery efforts.
- **Documentation and Investigation**: Drones capture high-resolution imagery and videos that can be used for post-incident analysis, insurance claims, and investigations into the cause of the fire.

The use of drones – which are equipped with advanced autonomy capabilities – in firefighting and other such disaster management scenarios enhances the effectiveness of response and rescue operations and also improves personnel safety.

### 2) Precision Agriculture

The role of drones in precision agriculture is transformative, enabling farmers to make informed decisions, optimize resource usage, and increase productivity while minimizing the environmental footprint of their operations [79]. Precision agriculture involves using technology to gather precise information about crop conditions, soil variability, and other factors that influence farming decisions. UAVs, specifically drones, contribute to precision agriculture in the following multi-faceted ways:

- **Aerial Imaging and Mapping**: Drones equipped with high-resolution cameras capture aerial imagery of agricultural fields. This imagery can be processed to create detailed maps that show variations in crop health, moisture levels, and growth. These maps provide farmers with insights into the spatial distribution of issues like pests, diseases, and nutrient deficiencies.
- **Crop Monitoring and Management**: Drones enable frequent and efficient monitoring of crops throughout the growing season. Farmers can identify stress factors early, such as inadequate irrigation, nutrient imbalances, or pest infestations. This allows for timely interventions, optimizing resource usage and minimizing crop losses.
- **NDVI and Multispectral Imaging**: Drones can carry multispectral cameras that capture images in various wavelengths, including those beyond the visible spectrum. Normalized Difference Vegetation Index (NDVI) calculations are used to assess plant health by analyzing the reflection of different light frequencies. NDVI maps help identify areas of low vigor or stress.
- **Variable Rate Application**: By analyzing data collected from drone surveys and mapping, farmers can create prescription maps for variable rate application of inputs like fertilizers, pesticides, and water. Drones equipped with sprayers can precisely apply these inputs to specific areas, reducing waste and environmental impact.
- **Yield Estimation**: Using aerial imagery and 3D modeling, drones contribute to accurate yield estimation by analyzing the size, density, and health of crops. This information aids in forecasting harvest quantities and planning logistics.
- **Disease and Pest Detection**: Drones help in early detection of diseases and pest outbreaks. Their ability to cover large areas quickly allows farmers to spot issues before they spread extensively, enabling targeted treatment.
- **Soil Health Assessment**: Drones can be equipped with sensors to analyze soil properties, moisture content, and compaction levels. This information guides decisions about soil management and irrigation.
- **Land and Resource Management**: Drones assist in land assessment, identifying areas with erosion, drainage problems, or soil compaction. This information helps plan land management strategies.
- **Environmental Monitoring**: Drones aid in monitoring conservation efforts, tracking biodiversity, and assessing the impact of agricultural practices on the surrounding environment.

### 3) Forest Planting

Drones with advanced AI capabilities like aerial mapping can accomplish planting more effectively [80]. Their unique features are essential for reforestation efforts:

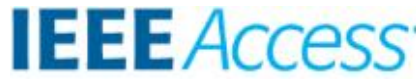

- **Access to Remote Areas**: Drones can reach areas that are difficult for humans to access, which can help speed up the reforestation process and ensure that trees are planted in the right locations.
- **Speed and Efficiency**: Drones can plant seeds at a much faster rate than manual planting. They can work together in a "swarm" to complete the task autonomously or with a single human supervisor overseeing the process.
- **Precision**: Drones can be equipped with specialized planting equipment, allowing them to quickly and accurately plant seeds directly into the ground. They can drop seeds along a predefined route.

Additionally, their ability to monitor and analyze the growth of newly planted trees can significantly contribute to the success of reforestation efforts, helping address the climate change problem more effectively.

## VI. MULTI-AGENT FLEETS – COORDINATED OPERATIONS

Multi-agent fleets of Urban Air Mobility (UAM) and other UAVs, including electric vertical takeoff and landing aircraft (eVTOLs) and delivery drones, will increasingly become indispensable for the modern cities with burgeoning population density. As urban congestion and traffic continue to escalate, the need for efficient, eco-friendly, and time-saving modes of transportation is paramount. Multi-agent UAM fleets can help alleviate this congestion and reduce the environmental impact of urban commuting by taking to the skies. However, to ensure the seamless integration of these fleets into our urban environments, it is crucial that they operate in a coordinated and economical manner. Similarly, for applications such as inspection and delivery, coordinated operations of drone fleets are undoubtedly required to ensure effective operations. However, as the number of AI-powered partial or fully autonomous UAS scales to thousands and beyond, technical and market mechanisms to operate the fleets optimally need to be in place.

### A. FLEET OPERATIONS AND MANAGEMENT

Integrating existing National Airspace System (NAS) operations with UAM operations faces several hurdles: 1) an increased volume of operations, 2) heightened operational density, 3) operations at lower altitudes, and 4) variations in the performance of different operators and air vehicles. These challenges place significant demands on the current air traffic control (ATC) system, indicating the shortcomings of the current Air Traffic Management (ATM) systems in effectively serving the large-scale UAM operations, and highlighting the necessity for substantial changes/amendments in this system [81]. Before we proceed, we need to define some terms and acronyms that are relevant to this topic:

- AGL - Above Ground Level. The altitude measured with respect to the underlying ground surface.
- MSL - Mean Sea Level. The average height of the surface of the sea for all stages of the tide.
- Flight Levels (FL). A measure of altitude (in hundreds of feet) used by aircraft flying above 18,000 feet with the altimeter set at 29.92 "Hg. For example, FL 200 means 20,000 feet MSL.
- ATM (right side of the figure) – UAS are certified and receive traditional air traffic services where required. This concept is based on the existing manned aviation system, where ATC provides separation and sequencing services to aircraft operating in controlled airspace. ATM requires UAS to comply with the same rules and regulations as manned aircraft, such as equipage, communication, navigation, surveillance, and identification.
- UTM (applicable up to 400 ft AGL) – UAS meet established performance requirements and cooperatively separate through shared situational awareness. Air Traffic Services (ATS) not provided. This concept is based on a distributed network of information providers and users, where UAS operators are responsible for planning, coordinating, and executing their own flights. UTM requires UAS to share their flight information and intentions with other UAS operators and stakeholders, such as local authorities, law enforcement, and emergency services.

**TABLE 3 Controlled and uncontrolled airspace classification in the U.S.**

| Class Name | Altitude Range | ATC requirements | Description |
|---|---|---|---|
| Class A | 18,000 - 60,000 ft | Mandatory ATC Clearance | High-altitude airspace |
| Class B | Around busy airports | Mandatory ATC Clearance | Busy airport airspace |
| Class C | Around medium-sized airports | Mandatory 2-way ATC Communications | Medium-sized airport airspace |
| Class D | Around smaller airports with ATC towers | Mandatory 2-way ATC Communications | Smaller airport with ATC tower |
| Class E | Around smaller airports without ATC towers | No mandatory ATC contact or clearance | Smaller airport without ATC tower |
| Class G[9] | Below 1,200 ft | No ATC Services, Visual Flight Rules | Uncontrolled airspace, VFR |

[9] Controlled airspace covers Classes A, B, C, D and E, while uncontrolled airspace covers Classes F and G. Class F airspace is not used in the U.S.

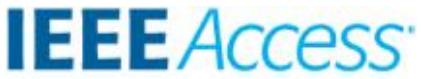

Table 3 provides a summary of controlled and uncontrolled airspace classification in the U.S. Class G airspace, also known as uncontrolled airspace, is the part of the airspace that isn't classified as Class A, B, C, D, or E. It is thus termed uncontrolled airspace. The extent of Class G airspace is from the surface up to the base of the Class E airspace that lies above it. Class G airspace is where the majority of the UAV applications are to be operated. While Air Traffic Control (ATC) doesn't have the authority or responsibility to manage air traffic in this space, there are minimums for Visual Flight Rules (VFR) that are applicable to Class G airspace. VFR are a set of regulations under which a pilot operates an aircraft in weather conditions generally clear enough to allow the pilot to see where the aircraft is going. Specifically, the weather must be better than basic VFR weather minima, i.e., in visual meteorological conditions (VMC), as specified in the rules of the relevant aviation authority. The pilot must be able to operate the aircraft with visual reference to the ground, and by visually avoiding obstructions and other aircraft. Visual Line of Sight (VLOS), on the other hand, refers to a type of UAS operation in which the aircraft is flown within the Pilot in Command's (PIC) visual line of sight. In essence, VLOS operations are similar to VFR in that they both require the operator to maintain visual contact with the aircraft. However, VLOS is specific to UAS, while VFR applies to all types of aircraft. Beyond Visual Line of Sight (BVLOS), conversely, are operations wherein the aircraft is flown beyond the PIC's or VO direct sight of the aircraft. BVLOS operations represent a departure from VFR because they do not require the operator to maintain visual contact with the aircraft. Instead, they rely on communication and sensing tools and technology such as the Remote Pilot Station (RPS) or Ground Control Station (GCS) for control. Fig 9 provides an integrated view of NAS and UTM operations where Class A through Class G vehicles operate with their respective VLOS and BVOLS protocols.

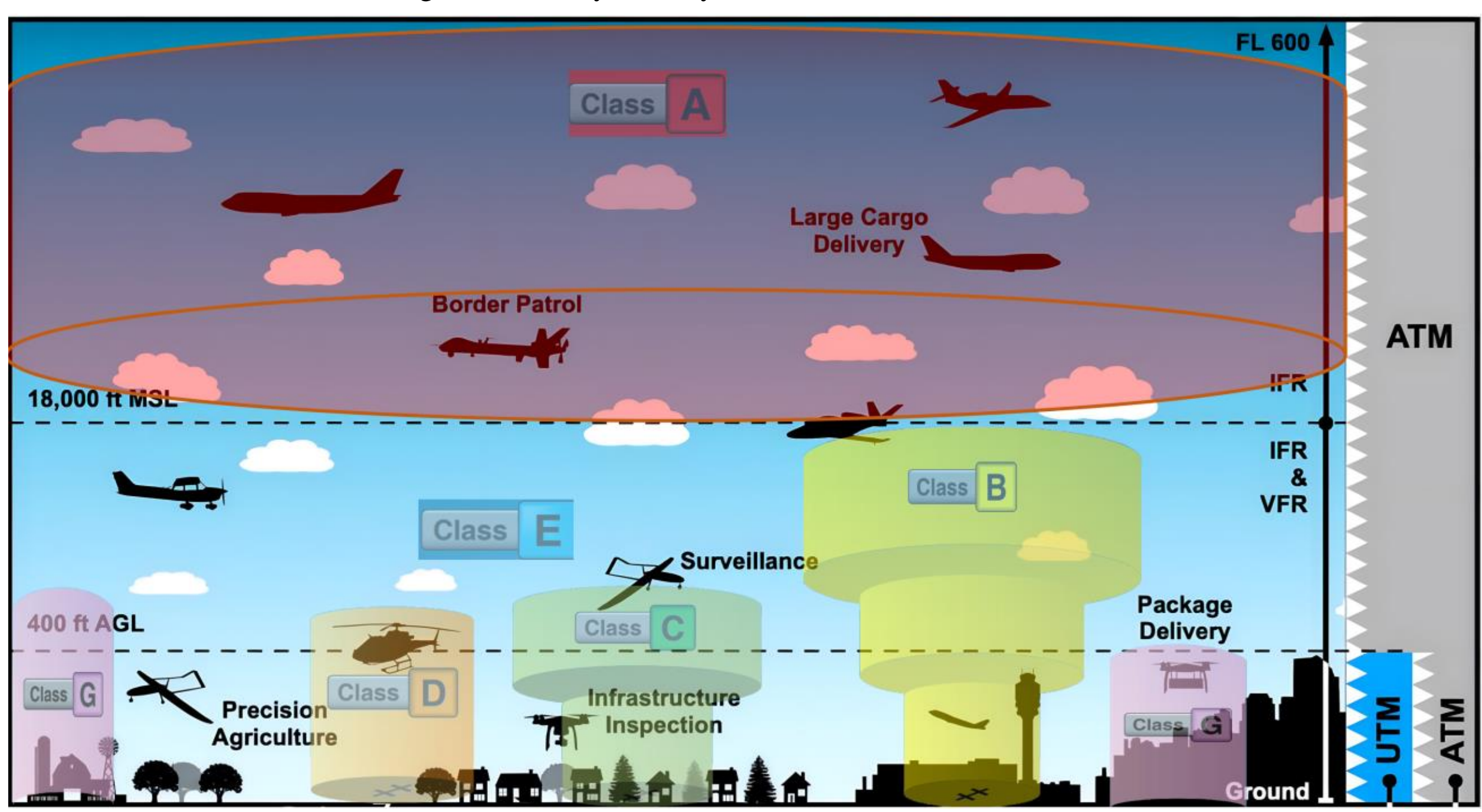

**FIGURE 9 Integration of National Airspace System (NAS) operations with Urban Air Mobility (UTM) operations. Airspace Classes A to G. Traditional and AAM vehicles sharing the airspace managed by ATM and UTM (see the right side of the diagram).**

1) UAS Traffic Management – Not mere expansion of ATM system

Unmanned Aircraft Systems (UAS) Traffic Management (UTM) Concept of Operations (ConOps) effort is undertaken collaboratively across FAA which includes participation from Air Traffic Organization (ATO), Office of NextGen (ANG), Aviation Safety (AVS) organization, and NASA. UTM is being developed as a scalable, flexible, and adaptable system that can support the full range of UAS operations and technologies, coexist with manned traffic, and minimize disruption to the existing ATM system. UTM design can also adapt to new technologies and automation. It is aimed at minimizing deployment and development time by using current industry-provided technologies and capabilities that meet performance requirements for safety, security, efficiency, environmental impacts, and privacy.

UTM is a system to support low-altitude UAS operations in the NAS. UTM system would integrate UAS operations in the airspace above buildings and below traditional aviation operations. It is developed by the FAA, NASA, and industry partners through research, testing, and standards development. UTM consists of a network of actors and services that exchange information and services to enable safe and efficient UAS operations. UTM provides a set of services to support UAS Operators in meeting regulatory and operational requirements. These include Performance Authorization, Airspace

Authorization, Operation Planning, Constraint Information & Advisories, Separation, and Remote Identification. UTM defines the roles and responsibilities of various actors and entities in the UTM ecosystem. The FAA establishes the regulatory framework and operational rules, while the Operators and USSs are responsible for the coordination and management of operations. UTM services include registration, airspace authorization, remote identification, de-confliction, weather, surveillance, and others. UTM presents five scenarios that demonstrate different aspects of UTM operations in uncontrolled and controlled airspace. These include nominal operations, UAS Volume Reservations, interactions with manned aircraft, remote identification, and public safety requests [82].

*a) UTM Airspace Management - A new operational paradigm*

Airspace management is a function of UTM that ensures UAS operations are authorized, safe, secure, and equitable in terms of airspace access. The factors impacting [83] the airspace design are summarized in Fig 10. UTM ConOps, introduced in the previous section, takes a layered approach to airspace management. It includes multiple layers including Performance Authorization, Airspace Authorization, Operation Planning, Constraint Information & Advisories, Separation, Remote Identification (RID), Contingency Management, Data Management and Access.

*Performance Authorization* is a process by which the FAA grants an Operator permission to conduct UTM operations based on their ability to meet performance criteria and requirements. *Airspace Authorization*, on the other hand, is a process by which the FAA grants an Operator access to operate in controlled airspace and provides situational awareness to air traffic facilities. *Operation Planning* is a process by which the Operator develops and shares their operation intent with other UTM participants and de-conflicts with other operations, airspace constraints, and environmental factors. *Constraint Information & Advisories* is a service by which the USS provides relevant data to the Operator, such as weather, terrain, obstacles, hazards, and *UAS Volume Reservations* (UVRs), to support safe and efficient UAS operations. *Separation* is a function by which the Operator maintains safe distance from other aircraft, airspace, weather, terrain, and hazards using shared intent, shared awareness, strategic de-confliction, vehicle tracking and conformance monitoring, and detect and avoid (DAA) technologies. *Remote Identification* (RID) is a function by which the UAS transmits a unique identifier and other information to enable identification of the UA/Operator by authorized entities and the general public. *Contingency Management* is another function by which UTM handles unexpected events or emergencies that may affect UAS operations, such as system failures, communication losses, weather changes, or airspace conflicts. Contingency management involves operation planning, coordinated procedures and response protocols, and pre-programmed system or vehicle responses to flight anomalies. *Data Management and Access* is a function by which UTM ensures the security, privacy, and integrity of data exchanged among UTM participants and stakeholders. Data management and access involves data protection measures, such as encryption and authentication; data access policies and controls; data archiving and retrieval; and data sharing agreements.

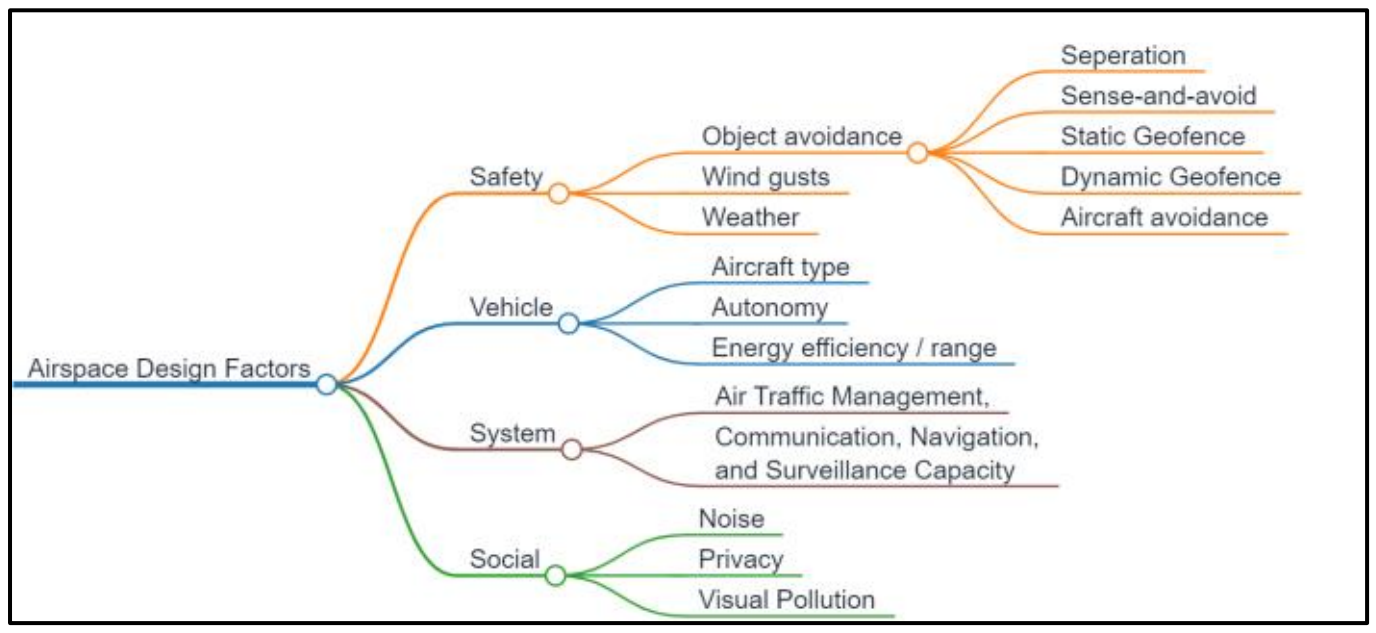

**FIGURE 10 Factors influencing the design of UTM operations.**

### 2) Vertiport Design, Placement, and Airspace Integration

Vertiport design refers to the physical layout and infrastructure of the landing and take-off sites for urban air vehicles. It is imperative that the design of vertiports considers their capacity to accommodate various vehicle types and sizes, in addition to accommodating diverse operational scenarios, including passenger transport, cargo delivery, and emergency services. Furthermore, it is essential for vertiports to incorporate communication, navigation, and surveillance systems to facilitate the secure and efficient execution of operations. Several critical factors impacting vertiport design encompass considerations of noise levels, safety measures, accessibility, environmental implications, and land utilization.

Vertiport placement pertains to the strategic positioning and dispersion of vertiports within urban environments, necessitating a careful evaluation of the demand and supply of urban air mobility services, as well as alignment with the pre-existing transportation infrastructure and land utilization patterns. Equally significant is the need for vertiport placement to mitigate adverse effects such as noise pollution, emissions, and visual disturbances on nearby communities. In addition, the coordination of vertiport placement with the urban airspace structure and traffic management system is of utmost importance to ensure the safe and efficient integration of urban air vehicles with other users of the airspace [84].

### 3) V2X communications

Vehicle-to-Everything (V2X) communication technology represents a transformative paradigm in the realm of intelligent transportation systems, where vehicles become active participants in a comprehensive network of information exchange [85]. V2X encompasses both Vehicle-to-Vehicle (V2V) and Vehicle-to-Infrastructure (V2I) communications, thereby enabling vehicles to interact with each other and infrastructure elements such as Air Traffic Control (ATS) systems, other UTM participants, and ground infrastructure. Leveraging wireless communication protocols, such as Dedicated Short-Range Communications (DSRC) or Cellular Vehicle-to-Everything (C-V2X), V2X empowers vehicles to share critical information, including real-time potential airspace conflicts, weather conditions, and even anticipate the movements of nearby vehicles [86].

### B. CHARGING THE BATTERY-POWERED UAM FLEETS

Deployment of fleets of UAMs underscores a critical imperative: the establishment of a comprehensive and sophisticated charging infrastructure. The majority of new designs and types of urban aerial vehicles are battery powered and hence it necessitates a nuanced approach, as their unique operational characteristics and requirements mandate specialized charging solutions. Urban environments, in particular, demand fast-charging networks, collocated with the vertiports and strategically positioned to accommodate UAM and UAS fleets [87]. Furthermore, the implementation of standardized charging protocols and interoperability standards assumes paramount importance to ensure seamless integration across a myriad of platforms.

#### 1) Charging Infrastructure

Charging infrastructure for UAVs like eVTOLs and drones, is a pivotal component that intricately intertwines the aerial mobility field with the broader energy grid. These vehicles require charging solutions that cater to their distinct operational needs, electric propulsion systems, and types of batteries installed on-board. The challenges are manifold, extending to the management of the low-voltage distribution grid, which must handle the increased demands posed by charging numerous electric aerial vehicles. To address this, charging infrastructure must be adaptable to the diverse battery charging needs of various UAVs, encompassing different charger designs and types, current and voltage ratings, and the time it takes to complete recharge. The standardization of charging protocols and interoperability standards is a critical step towards ensuring a seamless integration of these diverse platforms into urban environments where fast-charging networks are strategically positioned alongside vertiports to serve UAM and UAS fleets, enabling efficient, reliable, and sustainable aerial mobility solutions [88].

#### 2) Charging Station Operations

The operations of charging stations in the context of AAM demand a level of technical sophistication commensurate with the cutting-edge nature of urban air mobility (UAM) systems. These stations must employ optimized charging schedules and network routing algorithms to ensure the timely and efficient replenishment of electric aerial vehicle batteries [89]. The interplay between charging stations and on-demand mobility within both intercity and intracity UAM ecosystems requires precise orchestration to facilitate rapid, controlled, and conflict-free access to charging infrastructure. The debate over privatization versus public charging infrastructure will unfold in the next few years as various technological, capital, and regulatory forces play out. Yet, betting on an open-market-based system where various operators and service providers coexist, this coordination is vital for ensuring safety, minimizing airspace conflicts, and optimizing the utilization of airspace resources. These factors collectively represent the foundation upon which the seamless integration of electric aerial vehicles into urban environments hinges, fostering a safer, more efficient, and environmentally conscious aerial mobility landscape.

## VII. BENCHMARKING AND VALIDATION TO FACILITATE CERTIFICATION

"Standard" refers to the guidelines or requirements set by the regulators and industry, while "Certification" is the process of verifying adherence to these standards. Both play crucial roles in maintaining safety and quality in the aviation industry. Aviation is probably one of the most safety-conscious industries in modern times. Therefore, certification-readiness of any AI based technology that's built to be deployed in this industry is very tightly knit with the process of adhering to the safety-standards put forth by national and international agencies. These safety standards are collaboratively built by the regulatory bodies, industry bodies, and government. One of the challenges of developing and deploying UASs such, as eVTOLs and delivery drones, is obtaining the necessary certifications from the FAA in the U.S. (and respective regulatory agencies in other parts of the world) to operate them commercially. The FAA has a rigorous and complex process for ensuring the safety and reliability of aircraft, which involves three types of certifications: type, airworthiness, and production.

*Type certification* is for the aircraft design itself, which must meet certain performance and structural standards. Depending on the category of the aircraft, the FAA has different regulations that apply. For most UAS, drones, and multicopter eVTOLs, the relevant regulation is Title 14 Code of Federal Regulations, Part 21 [90]. For other eVTOLs, such as those that resemble conventional airplanes, the applicable regulation is Part 23 [91], which covers "normal, utility, acrobatic, and commuter category airplanes". *Airworthiness certification* is for the operation of a type-certified aircraft outside of the scope of Part 107, which governs small UAS operations. Airworthiness certificates can be either standard or special class, depending on the intended use of the aircraft. However, most UAS and eVTOLs do not qualify for a standard airworthiness certificate, as they do not meet the criteria established by the FAA. *Production certification* is for the manufacturing process of a type-certified aircraft, which must ensure consistent quality and conformity with the approved design. Production certificates are issued by the FAA Manufacturing Inspection District Offices (MIDO), while type certificates are issued by the FAA Aircraft Certification Offices (ACO).

Because most UAVs including eVTOLs are a new and innovative technology, the FAA does not have existing standards or regulations that fully address their unique features and capabilities. Therefore, the FAA relies on industry-developed standards, known as Means of Compliance (MOC), to evaluate and certify eVTOLs. The MOCs must be acceptable to the FAA and demonstrate how the UAV meets the performance criteria set by the agency. The certification process for UAVs begins with a proposal from the applicant, known as a G1 issue paper, which specifies the applicable standards and special conditions that must be met to achieve certification. The FAA reviews the proposal and either approves or rejects it. If it is rejected, the applicant must revise it to address the FAA's concerns and resubmit it. This process can take several iterations until a consensus is reached between the applicant and the FAA.

UAVs, built on a new conceptual foundation of autonomy, pose a new set of safety challenges, such as increased complexity, cyber-security threats, and human-machine interaction matters. The following subsections focus mainly on the software certification aspects as it pertains to the proposed Autonomy Blocks framework and their integration into the UAVs with certification requirements in mind.

### *A. SOFTWARE CERTIFICATION IN AVIATION*

DO-178C, "Software Considerations in Airborne Systems and Equipment Certification", is the primary document by which the certification authorities such as FAA, EASA and Transport Canada approve all commercial software-based aerospace systems. It is published by RTCA, Incorporated, in a joint effort with EUROCAE, and replaces DO-178B. It defines the software development process and the verification activities for each software level, from A (most critical) to E (least critical). It covers all aspects of software development, from planning and requirements to coding, testing, configuration management, and verification.

One of the challenges posed by the use of software in safety-critical applications is how to ensure confidence in the performance and behavior of complex systems that rely on machine learning and artificial intelligence. In particular, neural networks (NNs) are a type of machine learning technique that can learn from data and perform various tasks within the sensing, perception, planning, and controls modules. However, NNs are also difficult to understand, explain, and verify, due to their nonlinear and non-deterministic nature.

To address this challenge, EASA and Daedalean AG collaborated in an Innovation Partnership Contract (IPC) on the Concepts of Design Assurance for Neural Networks (CoDANN) . The purpose of this IPC was to investigate ways to gain confidence in the use of NNs in aviation, in the broader context of allowing machine learning and more generally artificial intelligence on-board aircraft for safety-critical applications. The project ran from June 2019 to February 2020 and resulted in a public report that presents the outcome of the collaboration.

#### 1) Means of Compliance with the Special Condition VTOL

In light of the dearth of appropriate certification criteria pertaining to the type certification of Vertical Take-off and Landing (VTOL) aircraft, an exhaustive set of specialized technical specifications has been meticulously formulated in the guise of a Special Condition for VTOL aircraft. This aforementioned Special Condition is explicitly tailored to cater to the unique attributes characterizing these aircraft, thereby prescribing airworthiness standards requisite for the conferment of a type certificate. These stringent requirements also encompass provisions for any modifications to an extant type certificate, all of which are directed towards person-carrying VTOL aircraft belonging to the small category. Notably, the purview of this Special Condition encompasses VTOL aircraft equipped with lift/thrust units designed for the generation of powered lift and control. In addition to these aforementioned aspects, the Special Condition VTOL simultaneously delineates safety and design objectives essential to this specialized realm [92].

#### 2) FAA / EASA Test Suite

The FAA conducted a joint research project with Daedalean, a company that develops machine learning applications for avionics, to study and flight test a neural network and vision-based runway landing guidance system for general aviation aircraft [93]. The system, called Visual Landing System (VLS), uses cameras and convolutional neural networks to detect and track runways and provide guidance cues to the pilot or the autopilot. The project aimed to evaluate whether the VLS can serve as a backup for other navigation systems in case of a GPS outage, and whether the W-shaped Learning Assurance process proposed by Daedalean can satisfy the FAA's intent for setting the future certification policy for machine learning systems. The W-shaped process consists of four main stages: data collection, model learning, model implementation and model verification. The flight test campaign took place in March 2021 in Florida, with FAA members on board, and involved 18 approaches over trained and untrained runways in various conditions. The results showed that the VLS performed well, detecting runways from up to 5 km away, and that the Learning Assurance process was compatible with the FAA regulatory framework.

EASA also collaborated with Daedalean on a series of joint studies to explore the key elements of the W-shaped development model for machine learning avionics software. The first study, Concepts of Design Assurance for Neural Networks (CoDANN) I [94], published in 2020, established the baseline understanding that the use of neural networks in safety-critical avionics applications is technically feasible. The second study, CoDANN II [29], published in 2021, focused on the implementation and inference parts of the W-shaped process, defined the role of explainability for the various actors involved in the certification process, and provided a system safety assessment process for integrating neural networks into avionics systems. The study also presented a case study of Daedalean's visual traffic detection system, which uses cameras and neural networks to detect cooperative and uncooperative traffic around the aircraft. The study demonstrated how the system can be verified using synthetic data, simulation and flight tests, and how explainability techniques can be used to analyze its performance and behavior.

In summary, both FAA and EASA have made significant progress in developing test suites for neural-network based software for aviation, in collaboration with Daedalean. These test suites are based on the W-shaped development model shown in Fig 11, which provides a structured and rigorous approach to ensure the safety and reliability of machine learning systems. The test suites also incorporate explainability methods, which enable the understanding and interpretation of neural network outputs and decisions. These test suites are expected to facilitate the certification of neural-network based software for aviation applications, such as runway landing guidance and visual traffic detection, especially for UAVs.

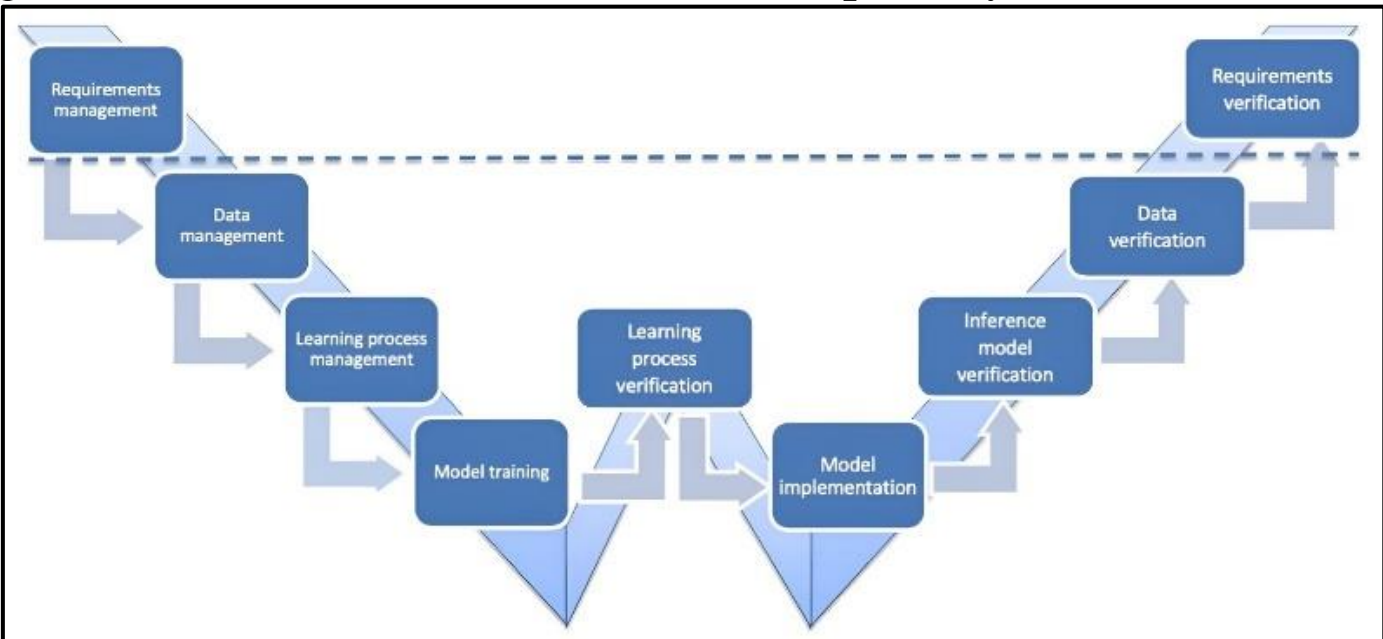

**FIGURE 11 W-shaped development cycle for Learning Assurance - Path to Certification of Neural Network based Autonomy Blocks**

### 3) NHTSA Test Suite for Ground-Air Package Delivery Drones

The National Highway Traffic Safety Administration (NHTSA) is the federal agency responsible for regulating the safety of motor vehicles and highway transportation in the United States. NHTSA also has a role in overseeing the integration of unmanned aircraft systems (UAS) or drones into the national airspace system, especially for ground-air package delivery operations. Ground-air package delivery drones are UAS that can transport goods from a ground vehicle to a customer's location using autonomous flight capabilities. These drones have the potential to improve the efficiency, convenience, and environmental impact of e-commerce and other delivery services.

However, ground-air package delivery drones also pose unique safety challenges that need to be addressed before they can be widely deployed. NHTSA has developed a test suite [95] for evaluating the performance and safety of these drones in various scenarios and environments. The test suite consists of a set of standardized procedures, metrics, and criteria that can be applied to different types of ground-air package delivery drones and operations. The test suite covers aspects such as:

- Ensuring the drone design and specifications meet the safety requirements and standards for UAS operations
- Testing and verification the drone flight control and navigation systems for accuracy, reliability, and robustness
- Development of secure the drone communication and data link systems from interference, jamming, or hacking
- Design and operations of the drone payload and delivery mechanisms to avoid damage, loss, or theft of goods
- Integration of the drone launch and recovery systems with the ground vehicles and infrastructure without causing traffic disruptions or accidents
- Compliance with traffic rules and regulations for both ground and air operations
- Obstacle and hazard detection and avoidance in the air and on the ground
- Respond to emergencies and contingencies such as weather, malfunctions, and collisions

The purpose of this test suite is to provide a consistent and objective framework for assessing the safety and performance of ground-air package delivery drones, which are likely equipped with varying autonomy capabilities, under various conditions and scenarios. The test suite can be used by drone manufacturers, operators, regulators, researchers, and other stakeholders to validate, verify, certify, or evaluate ground-air package delivery drones and operations. The test suite can also support the development of standards, best practices, and regulations for this emerging sector of UAS applications.

### 4) Challenges in adopting ISO 26262 / ASIL-D for Airborne Systems

ISO 26262 is an international standard that defines requirements and processes for ensuring functional safety of electrical and electronic systems in passenger vehicles [96]. Functional safety is the absence of unreasonable risk due to hazards caused by malfunctioning behavior of these systems. It defines guidelines to minimize the risk of accidents and ensure that automotive components perform their intended functions correctly and at the right time. ISO 26262 is based on the IEC 61508 standard for general industrial applications, but it is adapted to the specific needs and challenges of the automotive sector. ISO 26262 covers the entire lifecycle of safety-related automotive systems, from concept phase to development, production, operation, service, and decommissioning.

One of the main challenges for applying ISO 26262 to UAVs is the definition and classification of safety goals and automotive safety integrity levels (ASILs). Safety goals are high-level requirements that specify the necessary risk reduction for avoiding or mitigating hazards. Safety standards assign integrity levels to systems or functions based on initial consequences analysis, with clear guidance for integrity level identification. ASILs are a measure of the severity, exposure, and controllability of hazards, ranging from A (lowest) to D (highest). For example, an eVTOL system that controls the flight stability would likely have a high ASIL level, while a system that provides entertainment functions would have a low ASIL level. The allocated integrity level dictates the rigor and stringency of development processes. However, it is important to note that ISO 26262 provides guidance and examples for defining safety goals and ASILs for passenger vehicles, but not for UAVs / eVTOLs specifically. Fig 12 below shows the various integrity levels for DO-178C AND ISO 26262.

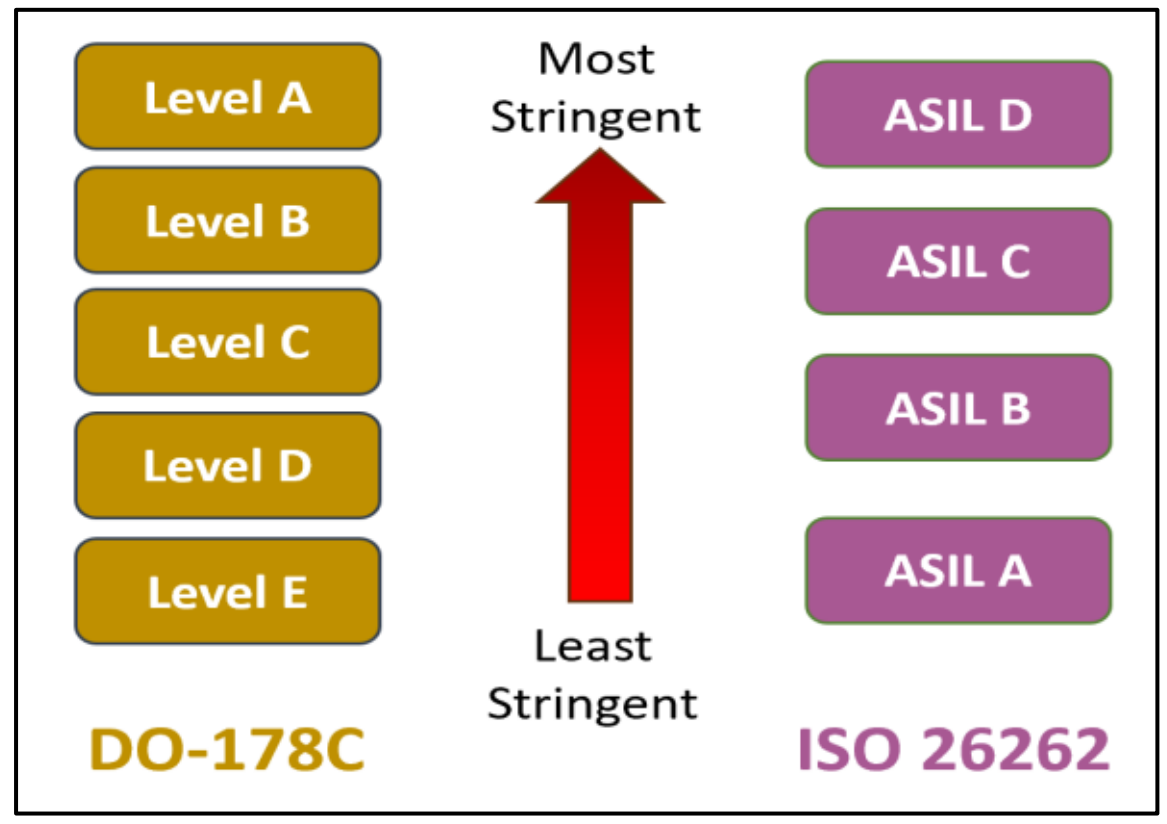

**FIGURE 12 DO-178C and ISO 26262 comparison.**

UAVs have different types of hazards and risks than passenger vehicles, depending on their size, weight, speed, payload, operation mode, mission type, flight environment, and regulatory framework. For example, a small UAV flying over a rural area may have a lower risk of causing harm than a large UAV flying over an urban area. Therefore, it is necessary to adapt the ISO 26262 methodology for defining safety goals and ASILs for UAVs according to their specific characteristics and scenarios.

Another challenge for applying ISO 26262 to UAVs is the verification and validation of autonomous functions. ISO 26262 proposes model-in-the-loop (MIL), software-in-the-loop (SIL), and hardware-in-the-loop (HIL) simulation for conducting software safety requirements verification. All of these simulation processes can be applied towards the common goal of generating autonomous vehicle requirements. However, simulation alone may not be sufficient to ensure the safety and reliability of autonomous functions that involve complex interactions with dynamic and uncertain environments. Therefore, it is necessary to complement simulation with real-world testing and evaluation of autonomous functions in representative scenarios and conditions.

Despite these challenges, applying ISO 26262 to UAVs also offers some opportunities for improving their safety and quality. One of the opportunities is the reuse and adaptation of existing standards and best practices from the automotive domain. ISO 26262 provides a comprehensive framework for managing functional safety throughout the lifecycle of safety-related systems. It also provides detailed guidance and recommendations for performing various activities and tasks related to functional safety. Therefore, applying ISO 26262 to UAVs can help to establish a common terminology, methodology, and documentation for ensuring functional safety of UAVs and their autonomous operations. It can also help to leverage existing knowledge and experience from the automotive domain and benefit from the lessons learned and good practices developed by other industries.

Another opportunity for applying ISO 26262 to UAVs is the innovation and advancement of new technologies and solutions for autonomous operations. ISO 26262 encourages the use of state-of-the-art methods and tools for designing, developing, testing, and operating safety-related systems. It also supports the continuous improvement and optimization of functional safety processes and products. Therefore, applying ISO 26262 to UAVs can stimulate the research and development of new technologies and solutions that can enhance the capabilities, performance, and efficiency of autonomous functions. It can also foster the collaboration and integration of different disciplines and domains that are involved in the creation and operation of *autonomous systems*.

## VIII. EXPLORING ADJACENT TERRITORIES: A CLOSER LOOK AT TWO KEY ASPECTS

### A. SIM-TO-REAL ROBOTS AND SYSTEMS

Sim-to-real robots and systems are challenging to develop and deploy due to the gap between simulation and reality [97]. In striving for a streamlined workflow to transition seamlessly from simulation to real-world applications, the key to success lies in: A) the meticulous optimization of the runtime and inference architecture, catering specifically to the target hardware and the intricacies of the application domain and B) comprehensively addressing all edge cases during the process of developing autonomy blocks, ensuring the robustness of algorithm performance to adhere to the standards of the safety-critical aviation industry. DO-178C and CoDANN help with the latter.

A fundamental aspect of this optimization pertains to domain adaptation. *Domain adaptation* refers to the process of customizing the sensing and perception modules of the robot according to the specific application domain [98]. For example, different types of sensors may be required for indoor and outdoor environments, or for different weather conditions. Moreover, the sensor data may vary significantly from simulation to reality, requiring robust and adaptive models that can handle domain shifts. Various techniques such as data augmentation, domain randomization, and adversarial learning can be leveraged to train and test the AI models (powering various Autonomy Blocks) in diverse and realistic scenarios.

*Online fine-tuning* refers to the ability of the robot to adapt its behavior and decision-making modules based on the feedback from the environment and the user [99]. Online fine-tuning enables the autonomous system to improve its performance based on continuous learning and real-world experiences, thereby promoting enhanced autonomy and reliability. For example, the aerial vehicle may need to adaptively adjust its speed, trajectory, or navigation strategy according to the dynamic and uncertain situations it encounters in the real world, which it has or hasn't necessarily encountered during the simulation-based training and testing. Methods such as reinforcement learning, imitation learning, and active learning can be employed to enable online learning and improvement of the AI models in an interactive and data-efficient manner.

*IoT edge device deployments* refer to the implementation of autonomy solutions on power-efficient embedded AI computing devices that can be integrated with the aerial vehicle's hardware [100]. For example, the UAV may need to run its models on a low-power CPU or GPU that can fit within its SWaP-c constraints. Moreover, the V2X capability will require the vehicle to communicate with other devices or cloud services via wireless networks, requiring reliable and secure data transmission protocols. The models need to be optimized for edge deployment using techniques such as model compression, quantization, pruning, and distillation, as well as leveraging edge computing platforms such as Azure IoT Edge. Leveraging the potency of embedded AI computing devices, this paradigm facilitates the efficient and seamless integration of autonomous capabilities into resource-constrained UAV environments.

### B. MONOLITHIC DEEP LEARNING FOR AUTONOMOUS AERIAL VEHICLES: CHALLENGES AND OPPORTUNITIES

Monolithic deep learning algorithms typically refer to comprehensive, end-to-end machine learning models that handle multiple aspects of autonomous flight [101]. These can include tasks such as obstacle detection and avoidance, path planning, and navigation. Monolithic deep learning models are tightly coupled systems where all the layers work in a highly synchronized manner. These models are often seen as a single, centralized unit, which can make them easier to develop, test, and debug. These algorithms are "monolithic" in the sense that they are designed to handle multiple tasks within the same framework, rather than relying on separate models or systems for each task. This can lead to more efficient and coordinated behavior in autonomous aerial robots. Monolithic models emphasize tight integration and synchronization of all components. However, this can also make them less flexible and adaptable compared to more modular or distributed systems.

In recent years, the field of AI has witnessed a significant transformation, shifting from task-specific, narrow models to larger, more versatile monolithic neural networks. For instance, within the domain of natural language processing (NLP), models like GPT-4 have demonstrated an impressive array of capabilities, encompassing tasks such as text summarization, translation, and sentiment analysis. Concurrently, visual-language models have been gaining proficiency in a multitude of tasks, including object detection, image captioning, and even generative tasks like creating artwork. This progression implies the potential for a unified, generalized model to potentially replace numerous task-specific models, offering enhanced efficiency and a simpler system architecture. However, when transitioning from NLP to the realm of robotics, a number of complexities come to the forefront.

Firstly, there is a notable scarcity of data (see Section III.B for challenges with Synthetic Data creation process), as end-to-end foundation models necessitate extensive training data, and there is a limited availability of curated datasets for pre-training robots. Consequently, the emphasis shifts towards enhancing the intelligence of existing foundation models for each of the proposed Autonomy Blocks (namely, Sense, Perceive, Plan, Actuate) even when their original application differs from that of aerial autonomy. Additionally, in the field of robotics, the wide variability in actuators and control systems introduces an additional layer of intricacy. Each type of aerial vehicle, whether it's a quadcopter, eVTOL, traditional aircraft, or helicopter, possesses a distinct set of actuators and corresponding control systems. This makes the concept of generalization more challenging. Employing a monolithic neural network that directly maps sensor inputs to actuator outputs is therefore not scalable and also risks overlooking the wealth of existing research in control theory.

## IX. CONCLUSION AND OUTLOOK

The outlook of the Advanced Aerial Mobility (AAM) field is poised for transformative change, with an increasing recognition of the need for AAM solutions in both urban and rural contexts. Urban congestion and gridlock have become a ubiquitous problem, and AAM holds the potential to alleviate these issues by introducing unmanned aerial vehicles (UAVs) for passenger and cargo transportation. In parallel, the use of UAVs is already revolutionizing various industries, from agriculture and construction to healthcare and logistics. These aircraft provide cost-efficient and rapid solutions for tasks such as crop monitoring, site surveys, medical supply delivery, and last-mile logistics, offering a glimpse into the future where AAM will play a pivotal role in enhancing productivity and quality of life.

The research and development of fully autonomous aerial vehicles is advancing at an impressive pace, propelling the AAM field toward its full potential. In this paper, we have presented a comprehensive study of the autonomous aerial mobility field, consisting of four main components: simulation, data, autonomy, and multi-agent fleets. We have described the functionalities and technical underpinnings of each component and how they interact with each other to enable safe and efficient operations of aerial vehicles, particularly AAMs based UAVs, in complex urban environments. We have outlined key innovations as well as existing systems. The focal point of our work is the autonomy blocks framework. This modular AI-based approach aims to address the full spectrum of autonomy for advanced aerial mobility, from sensing and perception to planning and control. We have proposed a customizable, modular, and extensible design paradigm that allows for building autonomy stack for different levels of autonomy and different types of aerial vehicles. We have also reviewed the state-of-the-art research and technologies in various domains and sectors that are relevant to our framework, including deep learning algorithms that cater to specific modules of the proposed autonomy stack.

Furthermore, we have discussed the challenges and opportunities for benchmarking and validating our framework based on the up-and-coming standards, guidelines, and ConOps being established by regulatory bodies around the world. Autonomous aerial vehicles need to comply with the tight regulatory oversight that governs the aerial mobility industry as ensuring the safety of passengers, property, and infrastructure is of paramount importance. This requires high standards of safety, security, and reliability. Therefore, AAM requires a multidisciplinary effort that integrates cutting-edge research and development from various fields, such as aviation engineering, computer science, artificial intelligence, robotics, and human factors. We believe that our autonomy blocks framework offers a holistic and comprehensive approach to developing the underlying technology – rooted in the multidisciplinary foundations - to advancing the field of autonomous aerial mobility. We hope that our work will inspire further research and innovation in this exciting and important domain.